\documentclass[11pt]{article}

\usepackage[preprint]{acl}

\usepackage{booktabs}
\usepackage{pifont}
\usepackage{xcolor}
\usepackage{amssymb}
\usepackage{tabularx}
\usepackage{array}
\usepackage{makecell}
\usepackage{multirow}
\usepackage{amsmath}
\usepackage{url}
\usepackage{hyperref}

\usepackage{times}
\usepackage{latexsym}

\usepackage[T1]{fontenc}

\usepackage[utf8]{inputenc}

\usepackage{microtype}

\usepackage{inconsolata}

\usepackage{graphicx}

\definecolor{ao(english)}{rgb}{0.0, 0.5, 0.0}
\newcommand{\cmark}{\textcolor{ao(english)}{\large\ding{51}}} 
\newcommand{\xmark}{\textcolor{red}{\large\ding{55}}}       

\newcommand{\benchmark}{OMHBench}

\title{\benchmark: Benchmarking Balanced and Grounded \\ Omni-Modal Multi-Hop Reasoning}

\author{
\textbf{Seunghee Kim}$^{1}$,
\textbf{Ingyu Bang}$^{1}$,
\textbf{Seokgyu Jang}$^{1}$,
\textbf{Changhyeon Kim}$^{1}$, \\
\textbf{Sanghwan Bae}$^{2}$,
\textbf{Jihun Choi}$^{3}$\thanks{This work was conducted at Sony AI.},
\textbf{Richeng Xuan}$^{4}$,
\textbf{Taeuk Kim}$^{1}$\thanks{Corresponding author} \\
$^{1}$Hanyang University,
$^{2}$NAVER Cloud,
$^{3}$Knowledge Work Inc., \\
$^{4}$Beijing Academy of Artificial Intelligence \\
\texttt{\{gyg9325, ingyu1008, diamondgyu, livex, kimtaeuk\}@hanyang.ac.kr,}\\
\texttt{baaesh10@gmail.com, jihun.choi@knowledgework.com, rcxuan@baai.ac.cn}
}

\begin{document}
\maketitle
\begin{abstract}

Multimodal Large Language Models (MLLMs) have increasingly supported omni-modal processing across text, vision, and speech.
However, existing evaluation frameworks for such models suffer from critical limitations, including modality shortcuts and biased reasoning paths.
To address these challenges, we propose \textbf{\benchmark}, a novel benchmark designed to rigorously evaluate omni-modal multi-hop reasoning.
It consists of 6,144 questions with balanced reasoning paths that are jointly grounded across all three modalities.
Extensive evaluation of 13 state-of-the-art models reveals that (1) a large performance gap exists between proprietary and open-source MLLMs and (2) even proprietary models exhibit high sensitivity to reasoning path variations, resulting in \textbf{asymmetric omni-modal grounding}.
Notably, models struggle when processing the speech modality, underscoring the need for balanced, multi-hop evaluation of omni-modal intelligence.

\end{abstract}

\section{Introduction}
\label{intro} 

Human perception and understanding are inherently complex, often requiring the integration of textual, visual, and auditory information.
Accordingly, the ability to process such heterogeneous inputs in tandem is fundamental to achieving human-level AI \cite{8269806}.
Relatedly, recent Multimodal Large Language Models (MLLMs) have evolved from initial bi-modal variants (e.g., text–vision and text–audio) to more comprehensive ones that jointly process text, vision, and audio, often referred to as \textbf{omni-modal} \cite{microsoft2025phi4minitechnicalreportcompact, xu2025qwen2, geminiteam2025geminifamilyhighlycapable}.\footnote{\textbf{Omni-modal} refers to the text-vision-audio setting, while \textbf{multi-modal} is a general term for more than one modality.}

The development of these models has also driven the emergence of new evaluation schemes, which fall into two main directions:
\textbf{Omni-Modal Understanding (OMU}) \cite{li2025omnibenchfutureuniversalomnilanguage, hong2025worldsense, zhou2025daily, chen2025uno, nguyen2025see}, which emphasizes measuring a model’s ability to collectively handle text, vision, and audio; and
\textbf{Cross-Modal Multi-Hop Reasoning (CMR)} \cite{talmor2021multimodalqacomplexquestionanswering, reddy2022mumuqamultimediamultihopnews, kim2025fcmr, foroutan2025wikimixqa, jang2025ict}, which focuses on its capability to perform multi-hop reasoning by composing information across modalities, typically in bi-modal settings.
The key distinction between the two lies in whether the speech modality is incorporated and whether multi-hop reasoning is explicitly required.

\begin{figure}[t]  
\centering
\includegraphics[width=1\columnwidth, keepaspectratio]{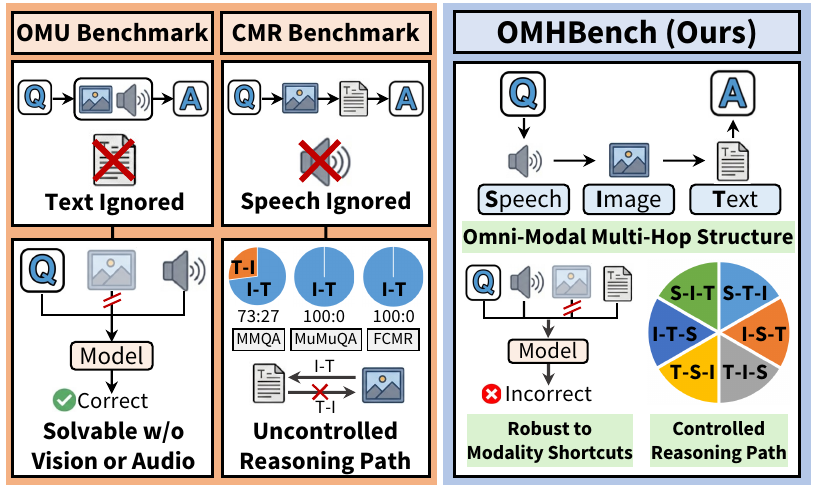}
  \caption{
Omni-Modal Understanding (OMU) benchmarks lack textual context and suffer from modality shortcuts, while Cross-Modal Multi-Hop Reasoning (CMR) datasets exclude speech and exhibit imbalanced reasoning paths. \benchmark\ addresses these issues.
}
\label{fig:previous_bench_vs_ours}
\end{figure}

In this work, we pose two crucial research questions regarding the current evaluation paradigms (see Figure~\ref{fig:previous_bench_vs_ours}):
(1) If an OMU benchmark can be solved without leveraging all three modalities, can it truly be said to evaluate omni-modal understanding?
(2) If a CMR benchmark is dominated by a single reasoning path, resulting in a heavily skewed composition distribution, can its results reliably reflect a model’s reasoning ability?
We demonstrate through experiments that both suspected pitfalls are present in practice and substantially undermine the integrity of current omni-modal evaluations.

\begin{table}[t]
\centering
\small
\renewcommand{\arraystretch}{0.9}
\setlength{\tabcolsep}{1.3pt}
\begin{tabular}{lcccccc}
\toprule
\multirow{2}{*}{\textbf{Benchmark}} &
\multicolumn{2}{c}{\textbf{Text}} &
\multirow{2}{*}{\textbf{Vision}} &
\multirow{2}{*}{\textbf{Speech}} &
\multirow{2}{*}{\textbf{CMR}} &
\textbf{Path} \\[-0.4ex]  
& \textbf{(Q)} & \textbf{(C)} & & & & \multicolumn{1}{c}{\textbf{Balance}} \\
\midrule
OmniBench       & \cmark & \xmark & \cmark & \cmark & \xmark & -- \\
WorldSense      & \cmark & \xmark & \cmark & \cmark & \xmark & -- \\
Daily-Omni      & \cmark & \xmark & \cmark & \cmark & \xmark & -- \\
OmniVideoBench  & \cmark & \xmark & \cmark & \cmark & \xmark & -- \\
UNO-Bench        & \cmark & \xmark & \cmark & \cmark & \xmark & -- \\
AV-SpeakerBench & \cmark & \xmark & \cmark & \cmark & \xmark & -- \\
\midrule
MMQA        & \cmark & \cmark & \cmark & \xmark & \cmark & \xmark \\
MuMuQA      & \cmark & \cmark & \cmark & \xmark & \cmark & \xmark \\
FCMR        & \cmark & \cmark & \cmark & \xmark & \cmark & \xmark \\
ICT-QA      & \cmark & \cmark & \cmark & \xmark & \cmark & \xmark \\
WikiMixQA   & \cmark & \cmark & \cmark & \xmark & \cmark & \xmark \\
\midrule
\textbf{\benchmark (Ours)} & \cmark & \cmark & \cmark & \cmark & \cmark & \cmark \\
\bottomrule
\end{tabular}
\caption{Comparison of OMU and CMR benchmarks by modality coverage and reasoning path balance. Text (Q) indicates question/option text, whereas Text (C) denotes separate contextual text. \benchmark\ uniquely supports all modalities with balanced multi-hop reasoning paths.}
\label{tab:related_work_comparison_benchmarks}
\end{table}

To systematically address these limitations, we propose \textbf{O}(mnimodal)\textbf{M}(ulti)\textbf{H}(op)\textbf{Bench}.\footnote{\benchmark\ is available at \url{https://huggingface.co/datasets/HYU-NLP/OMHBench}.}
By design, this novel benchmark departs from prior OMU and CMR datasets, as compared in Table \ref{tab:related_work_comparison_benchmarks}. 
It requires omni-modal multi-hop reasoning over text, image, and speech, with each modality explicitly used at least once, eliminating shortcuts that allow models to solve tasks without access to a specific modality.
Moreover, the ground-truth reasoning paths used as solutions are controlled with respect to modality order, enabling clearer identification of MLLMs' strengths and weaknesses.

Using \benchmark, we extensively test 13 proprietary and open-source MLLMs, uncovering several underexplored properties of omni-modal multi-hop reasoning.
We find that (1) entity-attribute-based multi-hop structure effectively mitigates modality shortcut issues; (2) model performance rankings can vary considerably depending on the category of reasoning paths; (3) models exhibit strong asymmetry in omni-modal grounding, especially when transferring semantics into the speech modality.

In sum, this work (1) exposes fundamental limitations of existing OMU and CMR benchmarks; (2) presents \benchmark, an omni-modal benchmark with controlled and balanced multi-hop reasoning paths; and (3) reveals systematic weaknesses in omni-modal grounding, particularly in speech.

\section{Related Work}
\label{related-work}

\paragraph{Multimodal Large Language Models (MLLMs)}
MLLMs aim to extend the capabilities of LLMs beyond text by incorporating additional modalities such as vision and audio.
Early efforts primarily focused on bi-modal settings, most notably text–vision \cite{alayrac2022flamingo,liu2023visualinstructiontuning} or text–audio \cite{kong2024audio,ghosh2024gama} integration, by attaching modality-specific encoders to off-the-shelf LLMs.
Consequently, these models remain limited in their ability to jointly reason over more than two modalities, motivating recent efforts toward omni-modal architectures \cite{xu2025qwen25omnitechnicalreport, yao2024minicpm, microsoft2025phi4minitechnicalreportcompact, ye2025omnivinci} that natively support text, vision, and audio within a unified framework.

\paragraph{Omni-Modal Understanding (OMU)}
With the advent of omni-modal models, a few benchmarks have been proposed to test their capabilities.
OmniBench \cite{li2025omnibenchfutureuniversalomnilanguage} is an initial attempt to evaluate models under tri-modal inputs.
WorldSense \cite{hong2025worldsense}, Daily-Omni \cite{zhou2025daily}, OmniVideoBench \cite{li2025omnivideobench}, and AV-SpeakerBench \cite{nguyen2025see} focus on vision-audio understanding.
UNO-Bench \cite{chen2025uno} further explores the relationship between uni-modal and omni-modal performance.

However, existing datasets emphasize visual and auditory signals, relegating text to questions or options, and often permit shortcuts that enable strong performance even without using all modalities.

\begin{figure}[t]  
\centering
\includegraphics[width=1.0\columnwidth, keepaspectratio]{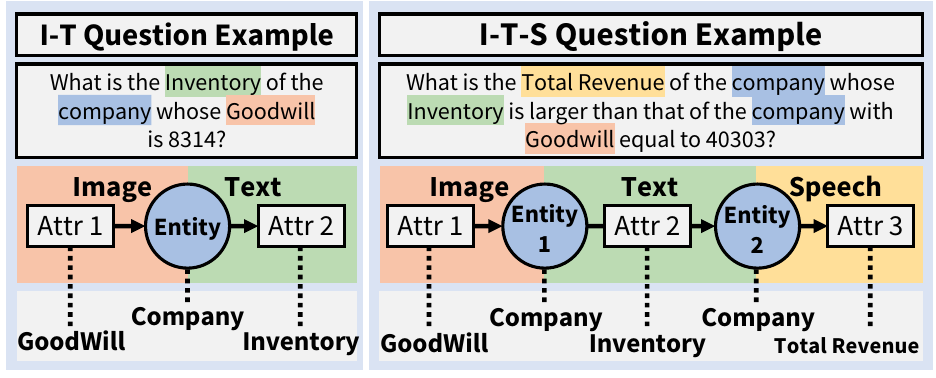}
\caption{Illustration of the proposed task formulation with two example questions and their reasoning paths: I-T (left) and I-T-S (right). Attributes (e.g., Goodwill) are accessible only through specific modalities, while entities (e.g., Company) are shared across modalities.}
\label{fig:formulation}
\end{figure}

\paragraph{Cross-Modal Multi-Hop Reasoning (CMR)}
CMR evaluates a model’s ability to perform multi-hop reasoning by interleaving textual and visual evidence.
Early benchmarks such as MMQA \cite{talmor2021multimodalqacomplexquestionanswering} and MuMuQA \cite{reddy2022mumuqamultimediamultihopnews} require models to integrate information from text, tables, and images, while FCMR \cite{kim2025fcmr} extends this paradigm to the financial domain.
ICT-QA \cite{jang2025ict} and WikiMixQA \cite{foroutan2025wikimixqa} also explore multi-hop reasoning over structured sources, e.g., tables and charts.

Nevertheless, these benchmarks primarily focus on text-vision modalities and do not support audio-based reasoning. 
Moreover, the absence of explicit control over reasoning paths in these datasets often leads to heavily skewed reasoning pattern distributions, compromising the reliability of evaluation.

\section{Preliminaries}
\subsection{Task Formulation: Omni-Modal Multi-Hop Reasoning}
\label{task-formulation-section}
Here, we specify the scope and formal definition of \textbf{omni-modal multi-hop reasoning} considered in this work.
Multi-hop reasoning inherently operates over \textit{entities} and their \textit{attributes} as articulated in the context.
From an omni-modal---more specifically, cross-modal---perspective, we consider scenarios in which entities are shared across modalities while attributes remain modality-specific, as illustrated in Figure \ref{fig:formulation}.
Answering such questions requires consulting modalities in a particular order, determined by the availability of the referenced attributes. We refer to this modality order as a \textbf{reasoning path}.

\begin{figure}[t]  
\centering
\includegraphics[width=0.8\columnwidth, keepaspectratio]{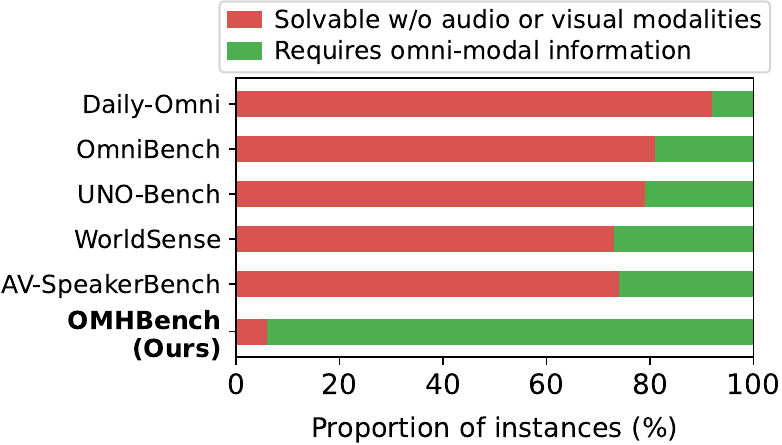}
  \caption{
Fraction of instances in the OMU benchmarks that remain solvable \textit{without} visual or auditory input. 
Across all datasets, nearly 70\textasciitilde80\% of instances are susceptible, indicating that shortcuts allowing models to answer without using certain modalities are prevalent.
}
\label{fig:previous_omni_bench_chart}
\end{figure}

For instance, the question \textit{``What is the inventory of the company whose goodwill is 8314?''} in Figure \ref{fig:formulation} necessitates integrating two attributes of the same entity, where \textit{goodwill} is available in the image modality and \textit{inventory} in text.
In this case, the reasoning path is I-T.
This formulation naturally generalizes to longer reasoning chains (e.g., I-T-S) through additional hops that identify subsequent entities.
A reasoning problem is considered \textit{omni-modal multi-hop} when the reasoning chain involves all three modalities: image, text, and speech.

\subsection{Shortcuts in OMU Benchmarks}
\label{previous-omu-shortcut}
To verify whether OMU benchmarks demand the exhaustive use of omni-modal input, we investigate five cases---OmniBench, WorldSense, Daily-Omni, UNO-Bench, and AV-SpeakerBench---by measuring the proportion of instances that remain solvable without visual or auditory input.
This experiment is conducted using Gemini 3 Flash \cite{gemini3Flash_modelcard}.

Figure \ref{fig:previous_omni_bench_chart} reports that nearly 70--80\% of instances in OMU benchmarks can be answered without access to certain modalities.
In other words, existing OMU benchmarks fail to genuinely assess the utilization of all three modalities due to insufficient structural constraints on cross-modal dependence.


\begin{figure}[t]  
\centering
\includegraphics[width=0.93\columnwidth, keepaspectratio]{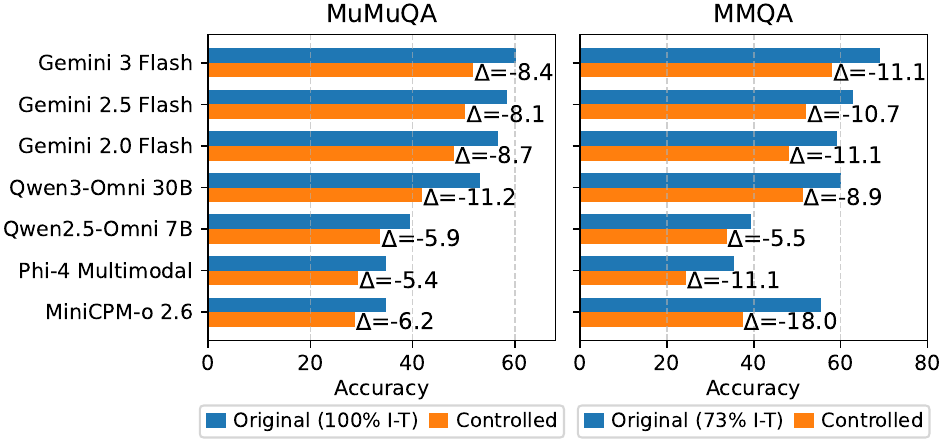}
  \caption{Performance comparison on the \textit{original} MuMuQA and MMQA datasets vs. their \textit{controlled} variants. When revised to ensure balanced reasoning paths, accuracy drops by up to 18\%, raising concerns about the validity of previous evaluations on the original datasets.}
\label{fig:CMR_IT_TI}
\end{figure}

\begin{figure*}[t]
\centering
\includegraphics[width=1.0\textwidth, keepaspectratio]{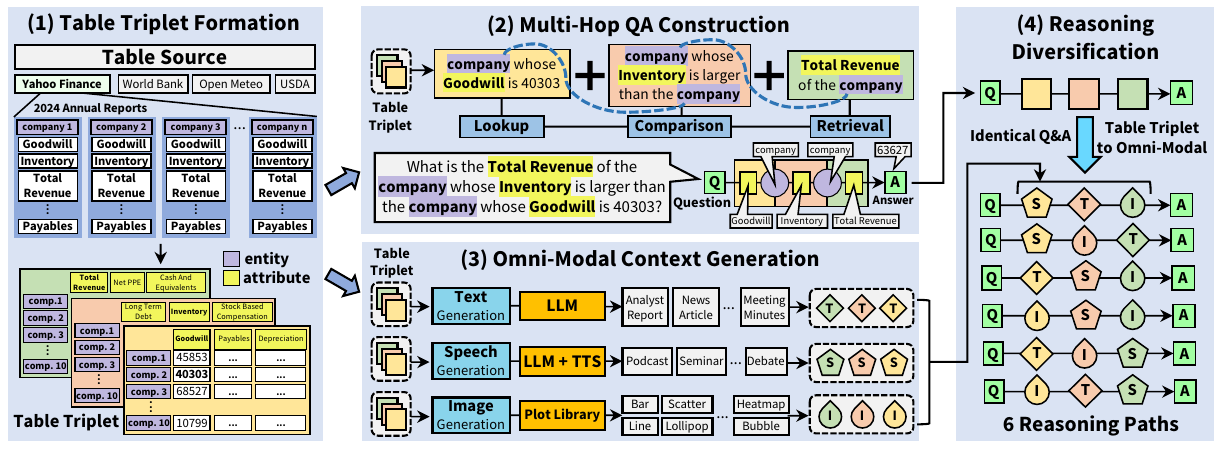}
  \caption{
Overview of the \benchmark\ pipeline.
(1) \textbf{Table Triplet Formation} constructs table triplets that share the same entities yet having separate attributes.
(2) \textbf{Multi-Hop QA Construction} yields multi-hop QA pairs by utilizing content from table triplets.
(3) \textbf{Omni-Modal Context Generation} converts each table into text, image, and speech modalities.
(4) \textbf{Reasoning Diversification} guarantees multiple reasoning paths via modality permutation.
}
\label{fig:dataset_generation_framework}
\end{figure*}

\subsection{Biases in CMR Benchmarks}
\label{previous-cmr-shortcut}
CMR datasets, e.g., MuMuQA and MMQA, are characterized by skewed distributions in their representation of reasoning paths.
As they consider only visual and textual inputs, there exist two possible orders: I-T and T-I.
Nonetheless, these benchmarks are imbalanced: MuMuQA and FCMR contain only I-T instances, while MMQA is skewed toward I-T, with roughly twice as many I-T as T-I.\footnote{ICT-QA \cite{jang2025ict} and WikiMixQA \cite{foroutan2025wikimixqa} are not publicly available, but their dataset construction does not consider reasoning path distributions.} 

To examine the effect of these biases, we conduct a controlled experiment by reversing the reasoning path direction (e.g., from I-T to T-I) for each question without changing its semantics.
This yields variants with uniform reasoning path distributions while preserving the originals’ unique properties. 
Specifically, we create balanced versions of MuMuQA, MMQA and FCMR, each with an equal split (50:50) between I-T and T-I instances.\footnote{The detailed process is described in Appendix \ref{sec:appendix_preliminary_cmr_dataset_gen}.}

Figure \ref{fig:CMR_IT_TI} shows that performance exhibits a pronounced and consistent decline under controlled settings, in some cases exceeding a 10\% drop in accuracy (see Figure~\ref{fig:CMR_IT_TI_FCMR} in the Appendix for FCMR).
This indicates that balanced reasoning path distributions are essential for accurate MLLM evaluation; however, simple path reversal offers only a partial remedy confined to bi-modal settings, highlighting the need for a new, systematic benchmark.

\section{Proposed Benchmark: \benchmark}
\label{dataset-construction}
Motivated by the largely isolated research streams on OMU and CMR and their respective limitations, we propose \textbf{\benchmark}, a new benchmark that bridges the two paradigms and addresses the aforementioned issues.
It enables controlled evaluation of omni-modal multi-hop reasoning with three key desiderata: (1) it prevents shortcut-prone evaluation by enforcing multi-hop reasoning, (2) it jointly incorporates textual, visual, and speech modalities, and (3) it explicitly controls reasoning paths to support reliable and unbiased assessment.

As shown in Figure \ref{fig:dataset_generation_framework}, \benchmark\ is constructed through four stages: (1) Table Triplet Formation, (2) Multi-Hop QA Construction, (3) Omni-Modal Context Generation, and (4) Reasoning Diversification.
We refer readers to Appendix \ref{sec:appendix_dataset_construction_detail} for details.


\subsection{Table Triplet Formation}
\label{subsec:Table Triplet Formation}
\benchmark\ covers four domains---finance, economics, climate, and nutrition---where reasoning over text, images, and speech naturally occurs, with instances evenly distributed across domains.
We employ real-world table data from Yahoo Finance, World Bank, Open-Meteo, and USDA.

Given an original tabular source, we construct table triplets of three smaller tables that share the same set of \textit{entities} but contain distinct \textit{attributes}. 
Each table has a size of 10 entities × 3 attributes and includes both relevant and distractor entities and attributes, requiring models to retrieve correct clues under information overload.

The intuition is that practical information is often organized in tabular form, yet its content can be realized in different modalities---images for visual comparison, text for detailed description, and speech for public announcements.
The table triplets serve as the core intermediate representation, which are used to construct question-answer pairs (\S\ref{table-triplet-mhqa-construction}) and corresponding omni-modal contexts (\S\ref{trimodal-generation}).



\subsection{Multi-Hop QA Construction}
\label{table-triplet-mhqa-construction}
From each table triplet, we formulate a question that requires three-hop reasoning to answer. 
In detail, we define eight reasoning operations---Lookup, Ranking, Comparison, Range, Proximity, Retrieval, Mean, and Summation---sample three of them, and sequentially apply each to the tables to derive questions.\footnote{In practice, not all combinations of reasoning operations are valid; their applicability depends on the entities and attributes involved. We therefore discard infeasible combinations and retain only those that apply to each case.}
The first two reasoning steps focus on entity-level reasoning, applying operations like Lookup, Comparison, and Range to select, compare, or filter entities and pass either a single entity or several entities forward. 
The final step produces the answer by either retrieving an attribute of a selected entity or aggregating attributes over a filtered entity set using operations such as Mean or Summation.

Note that the proposed procedure is deterministic and rule-based, enabling scalable question generation given a sufficient number of tabular data sources, without relying on costly external tools such as generative AI. Consequently, the QA construction process is efficient and fully automated.


Finally, we partition the QA pairs into two categories based on their underlying reasoning operation structures. \textbf{\benchmark-Connect} includes cases where intermediate results remain single entities throughout the reasoning process, following a fixed sequence of \textit{Lookup--Comparison--Retrieval} operations. 
By contrast, \textbf{\benchmark-Reasoning} covers cases where intermediate results expand into sets of entities, requiring aggregation operations. 

\subsection{Omni-Modal Context Generation}
\label{trimodal-generation}
At this stage, the three tabular sources from \S\ref{subsec:Table Triplet Formation} are transformed into contextual representations across three modalities---text, image, and speech.
We explain this part using the financial domain as an example; the same process can be applied to others.

For the text modality, we define scenarios such as analyst reports, news articles, and meeting minutes, following prior work in financial text mining \cite{kumar2016survey, pejic2019text, gupta2020comprehensive}.
Task-specific prompts are then used to guide LLMs in generating natural language descriptions grounded in the underlying tables.
For the image modality, we generate visualizations using plotting libraries, following common practices in financial data visualization \cite{ko2016survey, uddin2024data, christensen2024data}.
For the speech modality, we adopt the taxonomy of financial speech scenarios proposed by \citet{cao2025finaudio}.
The generated scripts are synthesized into speech using Kokoro-82M TTS \cite{hexgrad_2025}.
These are constructed as a multi-speaker dialogue in which each speaker describes different attributes of the same entity, encouraging models to leverage both semantic content and acoustic cues.
To enhance linguistic and stylistic diversity, we use three LLMs---GPT-5.1 \cite{openai_gpt5p1_systemcard}, Grok-4 \cite{xai_grok4_modelcard}, and Claude-Sonnet-4.5 \cite{anthropic_claude_sonnet45_systemcard}---for text and speech script generation.

\subsection{Reasoning Diversification}
\label{cmr-instance-construction}

We then pair a multi-hop QA instance (\S\ref{table-triplet-mhqa-construction}) with alternative configurations of omni-modal contexts (\S\ref{trimodal-generation}) to create multiple QA variants. 
They preserve the same question and answer, but differ in how contextual evidence is organized. 
By permuting modality assignments over the three tables, we obtain $3! = 6$ possible reasoning paths. 
Note that across these variants, the informational content remains unchanged; only the modality sequence required for inference (i.e., the reasoning path) varies.

\subsection{Quality Control}
Lastly, we apply four quality control methodologies to ensure dataset reliability and fair evaluation.

\paragraph{Entity Anonymization}
Given prior findings that CMR datasets may allow shortcuts via parametric knowledge \cite{kim2025fcmr}, we anonymize entity names with alphabetical codes (e.g., B, X), forcing models to rely solely on the provided context.

\paragraph{Consistency Checking}
We perform QA-based consistency checks \cite{fabbri2021qafacteval} by deriving factoid questions from the original tables (\S\ref{subsec:Table Triplet Formation}) and validating answers using the converted context (\S\ref{trimodal-generation}). 
We also apply a test where an LLM reconstructs the original tables from the converted modalities (\S\ref{trimodal-generation}) and compares them with the originals. 
Both checks achieve 100\% consistency, confirming the absence of factual loss or distortion.


\paragraph{Question Rephrasing}
To enhance linguistic diversity, we paraphrase questions using multiple LLMs: GPT-5.1, Grok-4, and Claude-Sonnet-4.5.
Paraphrasing quality is evaluated with the Lexical Deviation (LD) metric \cite{liu2022towards}, where our dataset achieves higher LD scores than the widely used PAWS dataset \cite{zhang2019paws} (0.32 vs. 0.13), indicating greater lexical diversity.

\paragraph{TTS Validation}
We evaluate TTS fidelity using ASR-based error rates (WER and CER) and speech quality metrics (STOI and SI-SDR), following \citet{kumar2023torchaudio}.
The results demonstrate high transcription accuracy and audio quality (WER: 0.03, CER: 0.02, STOI: 99.2, SI-SDR: 21.0).

\paragraph{Final Dataset Statistics}
\benchmark\ comprises 6,144 instances evenly distributed across six reasoning paths. 
The benchmark is divided into two subsets---\benchmark-Connect and \benchmark-Reasoning---each with 3,072 instances, based on the required reasoning operations.
Comprehensive dataset statistics, including diversity control across the three modalities, are reported in Table \ref{tab:dataset_statistics_table} of the Appendix.



\begin{table}[t]
  \centering
  \scriptsize
  \setlength{\tabcolsep}{2.0pt} 
  \renewcommand{\arraystretch}{1.1}
  \begin{tabularx}{\columnwidth}{
    @{}
    >{\raggedright\arraybackslash}X
    @{}
    *{6}{@{\hspace{1.2pt}}r@{\hspace{1.2pt}}}
    r
    r 
    @{}
  }
    \toprule
    \multirow{2}{*}[-0.9ex]{\textbf{Model}} &
    \multicolumn{6}{c}{\textbf{Accuracy by Reasoning Path (\%)}} &
    \multirow{2}{*}[-0.9ex]{\makecell{\textbf{Avg.}\\\textbf{(Acc.)}}} &
    \multirow{2}{*}[-0.9ex]{\textbf{PBS}} \\
    \cmidrule(lr){2-7}
    & \textbf{S-I-T} & \textbf{S-T-I} & \textbf{I-S-T} &
      \textbf{T-S-I} & \textbf{I-T-S} & \textbf{T-I-S} & & \\
    \midrule
    \textbf{\textit{Proprietary Models}} \\
    Gemini 3 Flash & 97.5 & 98.4 & 75.4 & 75.0 & 60.2 & 63.5 & 78.3 & 32.2 \\
    Gemini 2.5 Pro & 94.5 & 96.9 & 66.4 & 71.1 & 55.5 & 50.8 & 72.5 & 25.0 \\
    Gemini 2.5 Flash & 82.0 & 85.9 & 50.8 & 54.7 & 26.6 & 21.9 & 53.6 & 4.7 \\
    Gemini 2.5 Flash-lite   & 49.2 & 60.9 & 38.3 & 35.2 & 5.5 & 4.7 & 32.3 & 0.0 \\
    Gemini 2.0 Flash        & 28.9 & 33.6 & 26.6 & 29.7 & 4.7 & 6.2 & 21.6 & 0.0 \\
    Gemini 2.0 Flash-lite   & 35.9 & 32.8 & 21.1 & 11.7 & 2.3 & 2.3 & 17.7 & 0.0 \\
    \midrule
    \textbf{\textit{Open-Source Models}} \\ 
    Qwen3-Omni 30B  & 75.8 & 77.0 & 46.7 & 49.6 & 16.0 & 16.0 & 46.8 & 2.3 \\
    Phi-4 Multimodal & 26.6 & 23.6 & 21.5 & 18.4 & 0.6 & 0.0 & 15.1 & 0.0 \\
    Qwen2.5-Omni 7B         & 22.7 & 20.9 & 19.3 & 20.5 & 2.0 & 1.8 & 14.5 & 0.0 \\
    Qwen2.5-Omni 3B        & 12.7 & 17.6 & 15.6 & 14.6 & 1.2 & 2.0 & 10.6 & 0.0 \\
    OmniVinci                & 14.8 & 8.6 & 14.8 & 7.0 & 0.8 & 0.6 & 7.8 & 0.0 \\
    MiniCPM-o 2.6            & 8.0 & 10.9 & 7.4 & 8.4 & 1.2 & 0.2 & 6.0 & 0.0 \\
    Omni-AutoThink           & 7.6 & 6.6 & 8.0 & 6.1 & 0.6 & 0.0 & 4.8 & 0.0 \\
    \bottomrule
  \end{tabularx}

  \caption{
Accuracies and Path Balance Scores (PBSs) across six reasoning paths in \textbf{\benchmark-Connect}. 
Avg. denotes macro-averaged accuracy. 
PBSs (\S\ref{subsec:Path Balance Score (PBS)}) measure robustness to reasoning path variations.
}
  \label{tab:accuracy_by_reasoning_path_Connect}
\end{table}

\section{Experiments}

\subsection{Experimental Setup}
We evaluate 13 MLLMs in total: both proprietary models---Gemini series \cite{gemini2Flash_modelcard, comanici2025gemini, gemini3Flash_modelcard}---and open-source ones---Qwen3-Omni 30B \cite{xu2025qwen3omnitechnicalreport}, Phi-4-Multimodal \cite{microsoft2025phi4minitechnicalreportcompact}, Qwen2.5-Omni \cite{xu2025qwen25omnitechnicalreport}, OmniVinci \cite{ye2025omnivinci}, MiniCPM-o 2.6 \cite{yao2024minicpm}, and Omni-AutoThink \cite{yang2025omni}.\footnote{As of 2026-01-01, the Gemini series is the only proprietary model family that supports native omni-modal reasoning.}
For models that support explicit reasoning modes, we enable this capability by setting a thinking budget of 8,192 tokens.
All models are prompted using zero-shot chain-of-thought with a brief instruction and no fixed reasoning format.
Model outputs are parsed into discrete answers and scored as correct or incorrect.
We report \textit{exact match} accuracies across all six reasoning paths, along with the macro-average.
As \citet{tan2024order} shows that input modality order can affect model behavior---which we also observe in \S \ref{subsec:appendix_input_order}---we randomize the arrangement of omni-modal contexts to prevent such biases.

\begin{table}[t]
  \centering
  \scriptsize
  \setlength{\tabcolsep}{2.0pt} 
  \renewcommand{\arraystretch}{1.1}
  \begin{tabularx}{\columnwidth}{
    @{}
    >{\raggedright\arraybackslash}X
    @{}
    *{6}{@{\hspace{1.2pt}}r@{\hspace{1.2pt}}}
    r
    r 
    @{}
  }
    \toprule
    \multirow{2}{*}[-0.9ex]{\textbf{Model}} &
    \multicolumn{6}{c}{\textbf{Accuracy by Reasoning Path (\%)}} &
    \multirow{2}{*}[-0.9ex]{\makecell{\textbf{Avg.}\\\textbf{(Acc.)}}} &
    \multirow{2}{*}[-0.9ex]{\textbf{PBS}} \\
    \cmidrule(lr){2-7}
    & \textbf{S-I-T} & \textbf{S-T-I} & \textbf{I-S-T} &
      \textbf{T-S-I} & \textbf{I-T-S} & \textbf{T-I-S} & & \\
    \midrule
    \textbf{\textit{Proprietary Models}} \\ 
    Gemini 3 Flash & 55.9 & 58.8 & 49.8 & 49.6 & 40.0 & 42.6 & 49.4 & 8.6 \\
    Gemini 2.5 Pro & 53.9 & 51.6 & 52.3 & 47.7 & 41.4 & 46.1 & 48.8 & 10.9 \\
    Gemini 2.5 Flash & 32.0 & 30.5 & 17.2 & 24.2 & 10.9 & 10.9 & 21.0 & 0.0 \\
    Gemini 2.5 Flash-lite   & 18.8 & 21.1 & 15.6 & 8.6  & 0.0  & 0.0  & 10.7 & 0.0 \\
    Gemini 2.0 Flash        & 4.7  & 11.7 & 4.7  & 6.2  & 0.8  & 0.0  & 4.7  & 0.0 \\
    Gemini 2.0 Flash-lite   & 3.9  & 5.5  & 3.9  & 2.3  & 0.8  & 0.0  & 2.7  & 0.0 \\

    \midrule
    \textbf{\textit{Open-Source Models}} \\ 
    Qwen3-Omni 30B & 27.3 & 28.5 & 14.1 & 14.6 & 2.7 & 2.7 & 15.0 & 0.0 \\
    Phi-4 Multimodal & 0.6 & 0.4 & 0.2 & 0.0 & 0.2 & 0.2 & 0.3 & 0.0 \\
    Qwen2.5-Omni 7B        & 0.4  & 1.0  & 1.0  & 0.6  & 0.2 & 1.2 & 0.7  & 0.0 \\
    Qwen2.5-Omni 3B        & 0.8  & 0.6  & 0.2  & 0.0  & 0.4 & 0.2 & 0.4  & 0.0 \\
    OmniVinci               & 0.6  & 0.2  & 0.2  & 0.4  & 0.0 & 0.0 & 0.2  & 0.0 \\
    MiniCPM-o 2.6           & 0.0  & 0.0  & 0.0  & 0.0  & 0.0 & 0.0 & 0.0  & 0.0 \\
    Omni-AutoThink          & 0.4  & 0.2  & 0.4  & 0.2  & 0.0 & 0.0 & 0.2  & 0.0 \\
    \bottomrule
  \end{tabularx}
  \caption{
Accuracies and Path Balance Scores (PBSs) across six reasoning paths in \textbf{\benchmark-Reasoning}. 
Avg. and PBSs are defined as in Table \ref{tab:accuracy_by_reasoning_path_Connect}.
}
  \label{tab:accuracy_by_reasoning_path_reasoning}
\end{table}



\begin{figure}[t]  
\centering
\includegraphics[width=1\columnwidth, keepaspectratio]{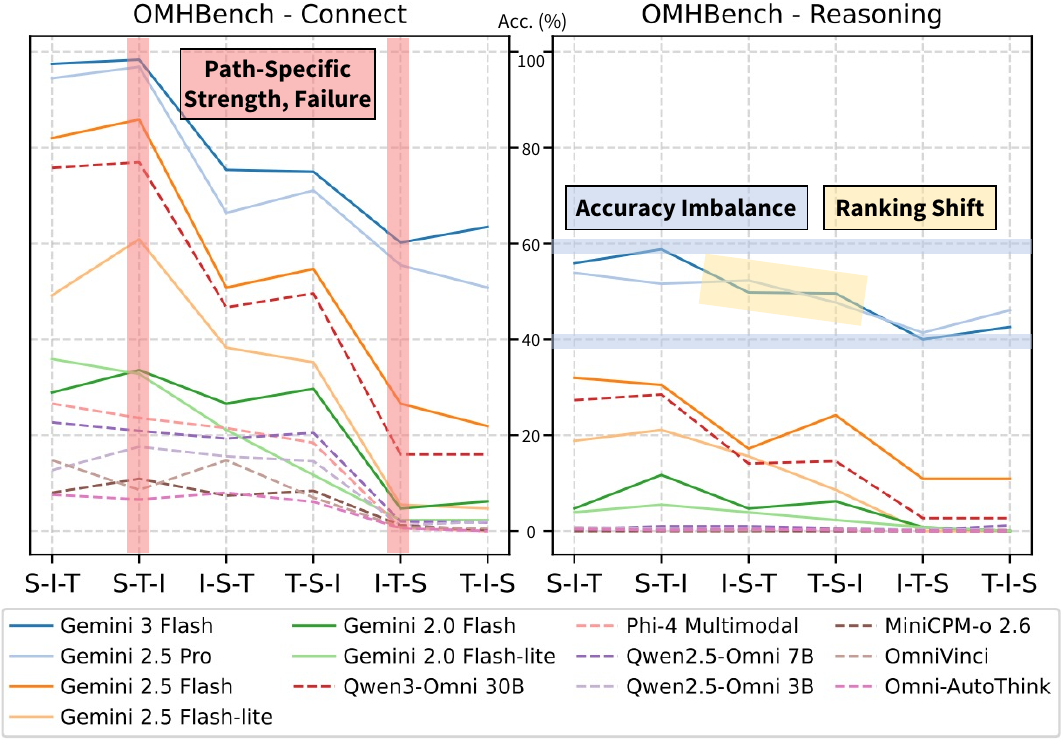}
  \caption{
Visualization of core trends from Table \ref{tab:accuracy_by_reasoning_path_Connect} and Table \ref{tab:accuracy_by_reasoning_path_reasoning}, highlighting \textcolor[HTML]{e57373}{path-specific strengths and failures}, \textcolor[HTML]{4f6fc2}{accuracy gaps}, and \textcolor[HTML]{ffb74d}{ranking changes by reasoning paths}.
}
\label{fig:result_fig}
\end{figure}

\subsection{Path Balance Score (PBS)}
\label{subsec:Path Balance Score (PBS)}
Beyond accuracy, we propose the \textbf{Path Balance Score (PBS)} as a novel metric to measure model robustness to variations in reasoning paths.
In \benchmark, each question is instantiated across all permutations of the available modalities; with $N$ modalities, this yields $N!$ reasoning paths sharing the same question.
PBS evaluates whether a model can consistently answer all such paths.

Formally, let the dataset contain $|D|$ instances, forming $|G| = |D|/N!$ groups.
For the $i$-th group, let $a_{i,j} \in \{0,1\}$ denote whether the model correctly answers the $j$-th path.
PBS is defined as:
\[
\scalebox{0.95}{$
\text{PBS} = \frac{1}{|G|}
\sum_{i=1}^{|G|}
\mathbb{I}\left(\sum_{j=1}^{N!} a_{i,j} = N!\right),
$}
\]
where $\mathbb{I}(\cdot)$ is the indicator function.
Intuitively, PBS counts a group as correct only if the model can answer the question under \textit{all} different reasoning paths, reflecting its robustness to path variations.

\subsection{Main Results}
\label{main-results-section}
Table \ref{tab:accuracy_by_reasoning_path_Connect} and Table \ref{tab:accuracy_by_reasoning_path_reasoning} report the performance of LLMs on \benchmark-Connect and -Reasoning, respectively.\footnote{For models allowing reasoning mode, only thinking variants are reported; full results are shown in Tables \ref{tab:accuracy_by_reasoning_path_Connect_full_model} and \ref{tab:accuracy_by_reasoning_path_Reasoning_full_model}.}
Figure \ref{fig:result_fig} provides a visual summary of these results and highlights several key trends.


\paragraph{Overall Trends}
For both \benchmark-Connect and \benchmark-Reasoning, proprietary models consistently outperform open-source models, with Qwen3-Omni 30B standing out as the strongest open-source model.
Importantly, we observe substantial performance variations across reasoning paths, with Qwen3-Omni 30B achieving 77\% on S-T-I versus 16\% on I-T-S in \benchmark-Connect.
\benchmark-Reasoning is more challenging than \benchmark-Connect, due to multi-entity intermediate states and numerical operations: even the best model reaches only 49.4\% accuracy, while most open-source models perform near zero.



\begin{figure}[t]  
\centering
\includegraphics[width=1\columnwidth, keepaspectratio]{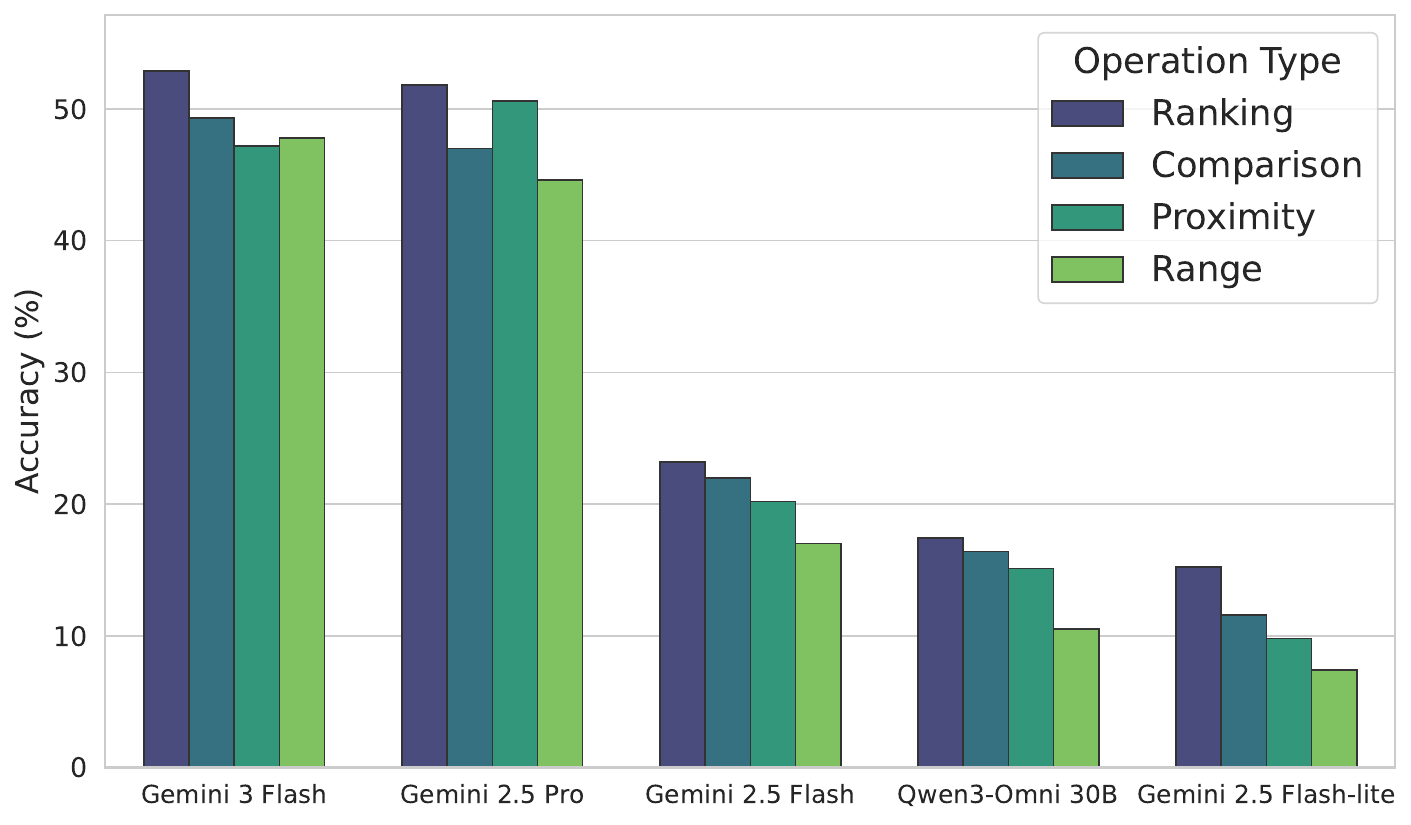}
  \caption{
Accuracy by operation type on \benchmark-Reasoning, computed over instances whose reasoning chains include the corresponding operation, showing decreasing performance from Ranking to Range.
}
\label{fig:reasoning-operation-chart}
\end{figure}

\paragraph{Difficulty by Reasoning Path}
We discover that MLLMs may answer the same question correctly or incorrectly depending on the composition of the omni-modal context. This behavior is clearly reflected by the low PBS scores observed for most models, indicating limited robustness to reasoning-path variations.
In particular, \textit{accuracy is strongly influenced by the position of the speech modality}: paths where speech appears earlier generally achieve higher performance, whereas those where speech appears later are substantially more challenging.
We further analyze this effect in \S \ref{step-by-step-analysis-section}.

In addition, Figure \ref{fig:result_fig} shows that model rankings can shift across different paths.
In \benchmark-Reasoning, Gemini 2.5 Pro outperforms Gemini 3 Flash on I-S-T, but underperforms on T-S-I.
This suggests that single-path evaluation (e.g., 100\% I-T in MuMuQA) fails to characterize model behavior, underscoring the limitation of existing benchmarks.

\paragraph{Difficulty by Reasoning Operation}
Instances in \benchmark-Reasoning are crafted to require diverse operations (e.g., Range, Ranking) for their solution.
To examine difficulty by operation type, we group instances accordingly and average accuracies over the top five models.
Figure \ref{fig:reasoning-operation-chart} highlights that performance degrades from Ranking to Comparison, Proximity, and Range, implying that MLLMs handle ordinal or pairwise comparisons well, while operations involving numerical neighborhoods or interval constraints remain challenging.

\begin{figure}[t]  
\centering
\includegraphics[width=1\columnwidth, keepaspectratio]{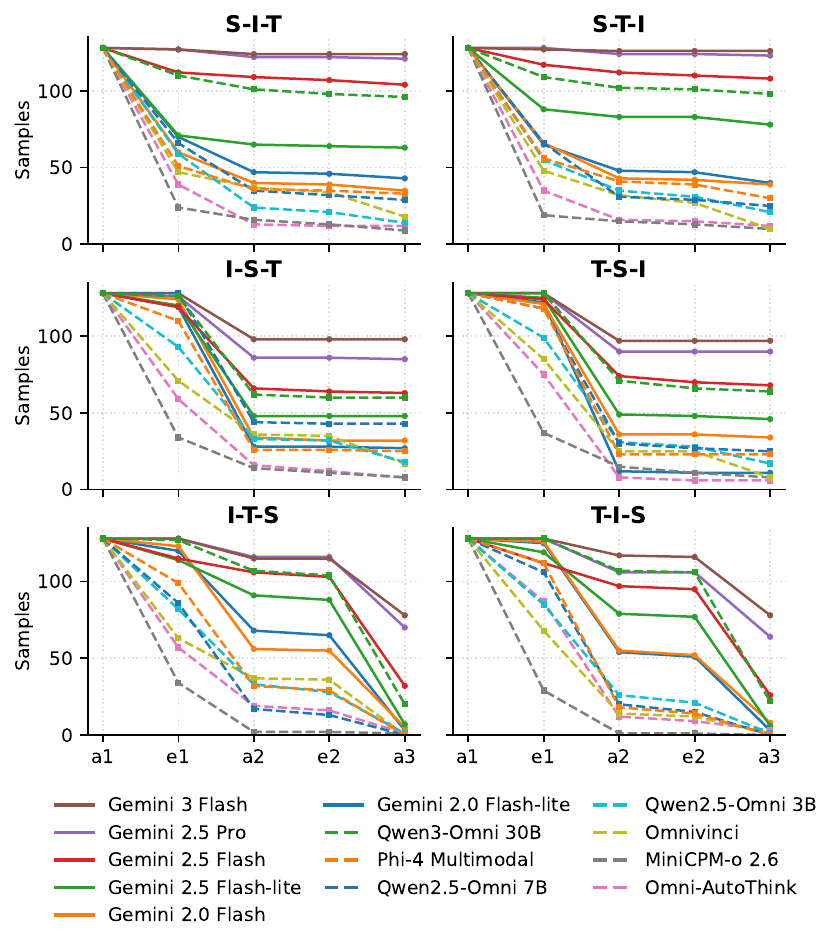}
  \caption{
Step-by-step failure analysis for six reasoning paths.
Each subplot reports the number of samples (out of 128 per reasoning path, 768 in total) successfully completing each reasoning stage---$a_{1}$, $e_{1}$, $a_{2}$, $e_{2}$, and $a_{3}$---with success depending on all prior steps being correct. 
This view reveals key bottlenecks in \benchmark.
}
\label{fig:stepbystep}
\end{figure}

\section{Analysis}

\subsection{Modality Shortcut Validation}
We verify that the modality shortcut issue observed in prior OMU benchmarks is no longer present in \benchmark.
Following the protocol in \S \ref{previous-omu-shortcut}, we measure the proportion of instances that remain solvable when one modality is removed.
Figure \ref{fig:previous_omni_bench_chart} confirms that \benchmark\ exhibits almost no shortcut-prone cases, reflecting the effectiveness of its explicit multi-hop design.
The few remaining solvable cases mainly stem from lookup-type questions, where correct answers can sometimes be obtained by chance through keyword retrieval.


\subsection{Step-by-Step Failure Analysis}
\label{step-by-step-analysis-section}
We conduct a step-by-step failure analysis to identify key bottlenecks in \benchmark, with results depicted in Figure \ref{fig:stepbystep}.\footnote{We perform this analysis on \benchmark-Connect for a controlled study of the required reasoning operations.}
Leveraging the task’s stepwise structure, we categorize failures by the stage at which the model fails to identify the required entity ($e$) or attribute ($a$), using Gemini 3 Flash.

Weaker models frequently fail at early reasoning stages, especially in identifying $e_1$ or $a_2$, regardless of the reasoning path, reflecting shortcomings in single-modal entity detection and cross-modal grounding.
In contrast, stronger models generally succeed in identifying $e_1$, but exhibit divergent performance at the $a_2$ stage depending on the reasoning path.
By decomposing each three-hop path into two cross-modal grounding steps, we find that transition between text and image (T-I and I-T), as well as from speech to other modalities (S-I and S-T), is relatively robust.
However, \textit{reasoning that moves to the speech modality (I-S and T-S) proves particularly challenging}.
We define this as \textbf{asymmetric omni-modal grounding}, underscoring inconsistencies in processing across modality orders.

\begin{figure}[t]  
\centering
\includegraphics[width=1\columnwidth, keepaspectratio]{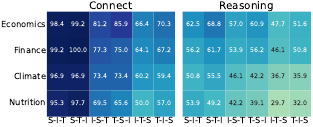}
  \caption{
Domain-wise accuracies of Gemini 3 Flash on \benchmark-Connect and -Reasoning, with rows denoting domains and columns denoting reasoning paths. Performance gaps are amplified for challenging paths.
}
\label{fig:domain_analysis}
\end{figure}

\subsection{Domain-Specific Analysis}
Figure \ref{fig:domain_analysis} presents the domain-specific performance of Gemini 3 Flash.
Performance varies across domains, with a maximum gap of 21.8\% between the economics and nutrition domains under T-S-I in \benchmark-Reasoning.
This implies that even the best model lacks uniform domain generalization, 
performing better on common domains (e.g., economics) than on technical ones (e.g., nutrition).

\begin{figure}[t]  
\centering
\includegraphics[width=1\columnwidth, keepaspectratio]{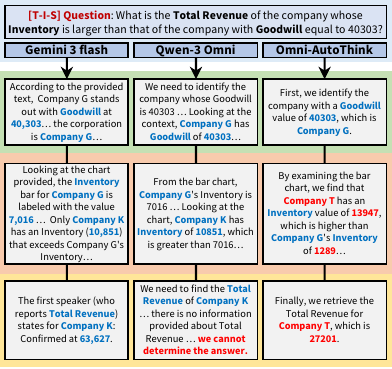}
  \caption{
Three case studies show (1) adherence to the intended multi-hop reasoning path without explicit guidance, (2) neglect of the speech modality depending on its position, (3) error accumulation in weaker models.
}
\label{fig:case_study}
\end{figure}

\subsection{Case Study}
To complement quantitative analyses, we present case studies in Figure \ref{fig:case_study} with three key findings.
(1) Even without explicit guidance in the prompt (i.e., zero-shot CoT), models consistently attempt to follow the intended reasoning path, confirming that \benchmark\ requires structured multi-hop reasoning.
(2) Models sometimes behave as if the speech modality were absent, reporting missing evidence despite speech information being provided; this behavior depends on the position of speech within the reasoning path (with similar input context lengths).
(3) Weaker models exhibit error accumulation, where early mistakes propagate and result in cascading failures at later reasoning stages.




\begin{figure}[t]  
\centering
\includegraphics[width=1.0\columnwidth, keepaspectratio]{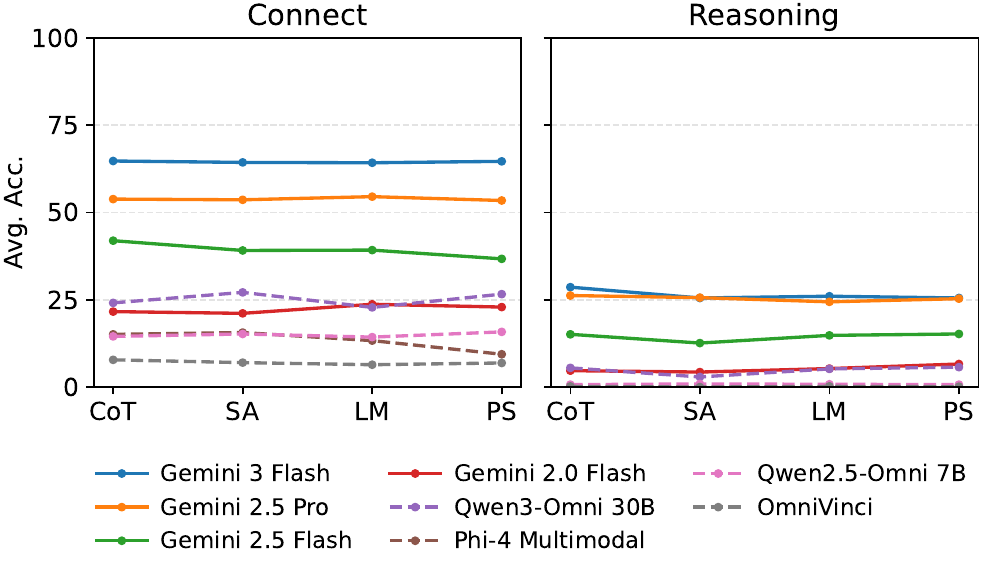}
  \caption{
Performance of four prompting methods on \benchmark: Chain-of-Thought (CoT), Self-Ask (SA), Least-to-Most (LM), and Plan-and-Solve (PS). They yield limited gains, calling for dedicated future research.
}
\label{fig:distinctprompt}
\end{figure}

\begin{figure*}[t]
\centering
\includegraphics[width=1.0\textwidth, keepaspectratio]{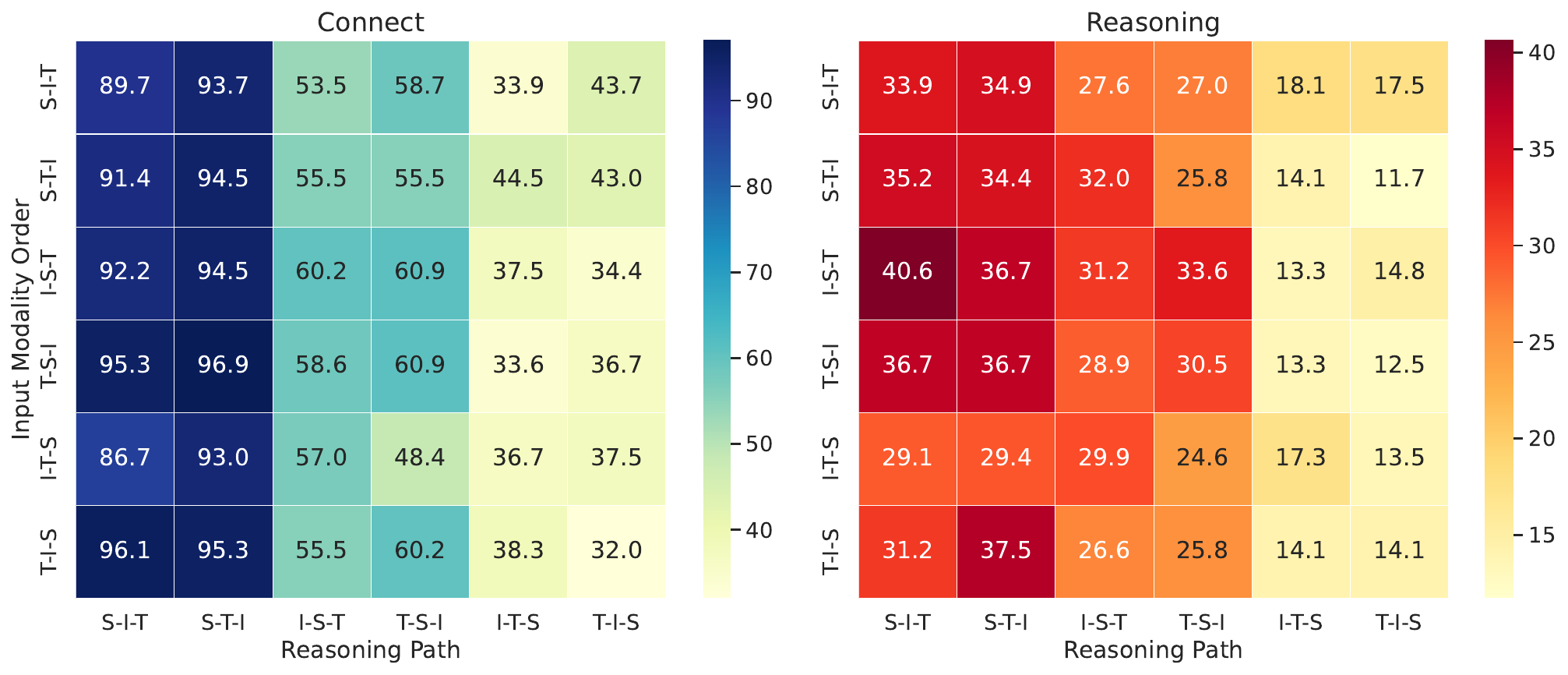}
  \caption{
Heatmaps of performance across all combinations of input modality order (rows) and reasoning path (columns) on \benchmark-Connect and \benchmark-Reasoning.
Input modality order refers to the sequence in which the three modal contexts (Speech, Image, Text) are presented to the model as input, while reasoning path refers to the modality that must be consulted at each hop. 
Each cell reports the exact match accuracy (\%) of Gemini 3 Flash under that configuration.
}
\label{fig:appendix_input_order_heatmap}
\end{figure*}

\subsection{Prompting Alone is Insufficient}
To examine whether the identified challenge can be alleviated by simply adopting advanced prompting techniques, we evaluate three multi-hop prompting strategies: Self-Ask \cite{press2023measuring}, Least-to-Most \cite{zhou2022least}, and Plan-and-Solve \cite{wang2023plan}.
As reported in Figure \ref{fig:distinctprompt}, none of these methods yields consistent improvements over the standard chain-of-thought baseline.
This suggests that \textit{asymmetric omni-modal grounding} is not primarily caused by insufficient prompt optimization, but instead reflects a fundamental limitation in transferring semantic representations across modalities, particularly into the speech modality.

\subsection{Impact of Input Modality Order}
\label{subsec:appendix_input_order}
Prior work has shown that altering the input order can affect model behavior \cite{chen2024premise, tan2024order}.
Moreover, models tend to prioritize information presented earlier in the input sequence, which can influence the reasoning process \cite{liu2023lost, wallace2024instruction}.
To examine the impact of input modality order---the sequence in which the three modal contexts are presented to the model, which is independent of the reasoning path---on performance, we conducted a systematic analysis on \benchmark\ using Gemini 3 Flash.
We enumerated all six possible input modality orders and evaluated accuracy across all 36 combinations of input orders and reasoning paths, as shown in Figure \ref{fig:appendix_input_order_heatmap}.
We observe that performance varies non-negligibly across input orders even when the reasoning path is held fixed.
For \benchmark-Connect, accuracy varies by up to 12.5 percentage points across input orders under the T-S-I reasoning path.
For \benchmark-Reasoning, the corresponding variation reaches 11.5 percentage points under S-I-T.
Although the overall trends across reasoning paths remain broadly consistent, these input-order-induced fluctuations can still introduce noise into comparative analysis.
Therefore, we randomize the input modality order in our main experiments to reduce this source of variance.



\section{Conclusion}
We present \benchmark, a dataset for robust evaluation of omni-modal multi-hop reasoning.
It addresses limitations of prior benchmarks by enforcing joint grounding across all three modalities and ensuring balanced proportions of distinct reasoning paths.
Experiments on \benchmark\ provide new insights into how MLLMs perform multi-hop reasoning, revealing that they are sensitive to modality orders and struggle particularly with cross-modal grounding.
In future work, we plan to explore training methods to improve the core multi-hop reasoning capabilities of omni-modal models.

\section*{Limitations}
\benchmark\ adopts an entity-attribute based formulation with fixed three-hop omni-modal reasoning chains to enable controlled and balanced evaluation of reasoning paths and to prevent modality shortcuts.
While this design facilitates precise analysis of modality interactions and fair comparison across different reasoning paths, it primarily targets reasoning scenarios that can be expressed through explicit entity-attribute relations and fixed-depth chains.
Extending the benchmark to support more diverse reasoning patterns is a promising direction for future work.

\section*{Acknowledgments}
This work was supported by Institute of Information \& communications Technology Planning \& Evaluation (IITP) grant funded by the Korea government(MSIT) (No.RS-2020-II201373, Artificial Intelligence Graduate School Program(Hanyang University)).
This work was supported by Institute of Information \& communications Technology Planning \& Evaluation (IITP) under the artificial intelligence semiconductor support program to nurture the best talents (IITP-(2026)-RS-2023-00253914) grant funded by the Korea government(MSIT).
This work was supported by the National Research Foundation of Korea(NRF) grant funded by the Korea government(MSIT) (RS-2025-00558151).

\bibliography{custom}

\clearpage
\appendix

\section{Preliminary CMR Dataset Construction}
\label{sec:appendix_preliminary_cmr_dataset_gen}
Since MMQA and MuMuQA are formulated as short-answer question answering tasks, while FCMR follows a multiple-choice format, we handle these datasets separately during the preliminary CMR dataset construction stage.

\begin{figure}[t]
\centering
\includegraphics[width=0.95\columnwidth, keepaspectratio]{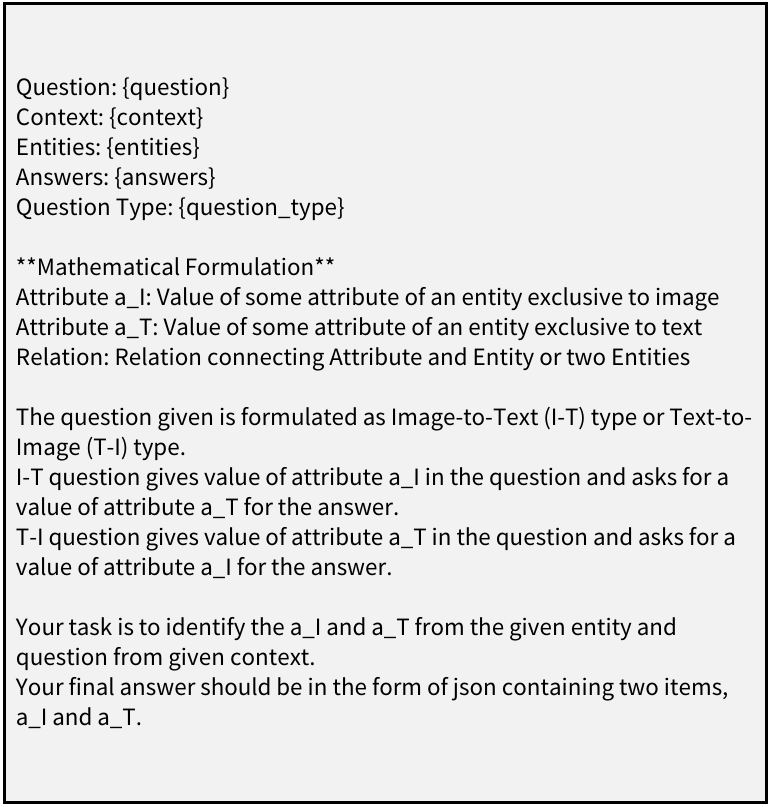}
  \caption{
Prompt for extracting attributes for a entity from each question in MMQA and MuMuQA.
}
\label{fig:appendix_entity_extraction_prompt}
\end{figure}

\begin{figure}[t]
\centering
\includegraphics[width=0.95\columnwidth, keepaspectratio]{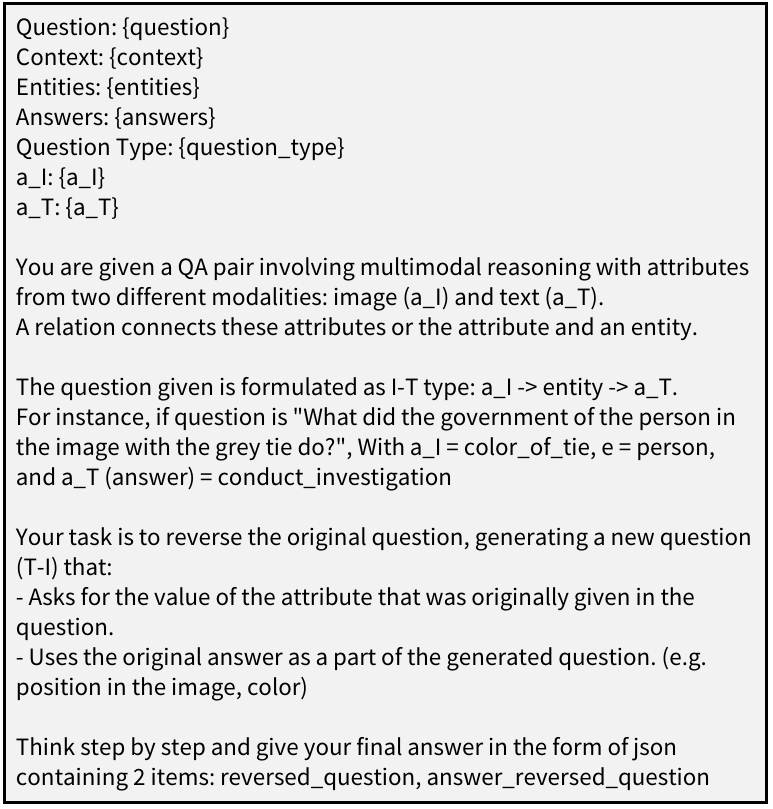}
  \caption{
Prompt for generating question from our formulation in MMQA and MuMUQA.
}
\label{fig:appendix_prev_cmr_question_gen}
\end{figure}

\begin{figure}[t]
\centering
\includegraphics[width=0.95\columnwidth, keepaspectratio]{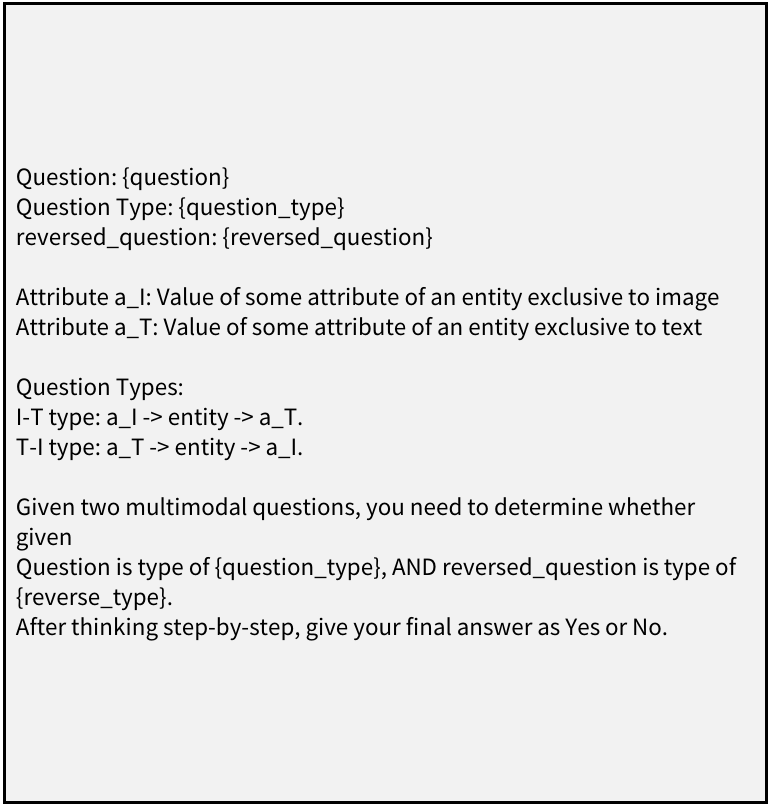}
  \caption{
Prompt for validating the reasoning order of generated questions in MMQA and MuMuQA.
}
\label{fig:appendix_prev_cmr_order_val}
\end{figure}

We first preprocess MMQA to align its format with that of MuMuQA, thereby simplifying the overall pipeline design. 
Specifically, we treat the table input in MMQA instances as a text modality and extend the context column of each instance to include a textual representation of the given table.
Next, we select instances from both MuMuQA and our preprocessed MMQA, that can be reformulated using our attribute-entity formulation.
This step is necessary because certain instances in MMQA do not require cross-modal reasoning and thus cannot be reversed.
We then prompt Gemini 2.0 Flash to decompose each question into its constituent entity and attributes, followed by question-generation step with the prompts provided in Figure \ref{fig:appendix_entity_extraction_prompt} and \ref{fig:appendix_prev_cmr_question_gen}.
Upon manual inspection of the generated questions, we find that some instances explicitly include the entity or the answer within the question text; we discard such cases.
Finally, we use an LLM-based validation step to further filter out questions with an incorrect reasoning order, using the prompt in Figure \ref{fig:appendix_prev_cmr_order_val}.
We pair up instances of two direction---I-T, and T-I---using the generated questions and their original counterparts.
Based on these pairs, we construct two sub-datasets: Original, which consists of instances from the original dataset, and Controlled, which includes both directional pairs. 

\begin{figure}[t]
\centering
\includegraphics[width=0.95\columnwidth, keepaspectratio]{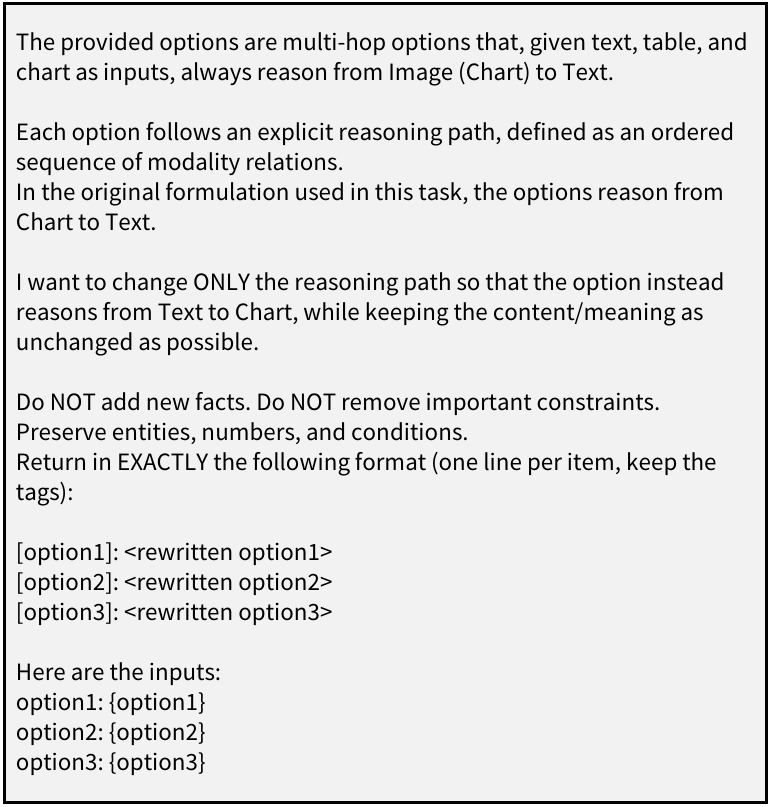}
  \caption{
Prompt for controlling the reasoning path of each option in FCMR.
}
\label{fig:appendix_prev_cmr_FCMR_Gen}
\end{figure}

For FCMR, the dataset construction process differs from MMQA and MuMuQA.
Since FCMR is designed with a template-based structure and explicit multi-hop reasoning paths, we directly apply Gemini 3 Flash to generate controlled instances over the answer options.
Using this procedure, we construct both Original and Controlled versions of the FCMR dataset.
The detailed generation prompt is provided in Figure \ref{fig:appendix_prev_cmr_FCMR_Gen}.


\begin{figure}[t]  
\centering
\includegraphics[width=1.0\columnwidth, keepaspectratio]{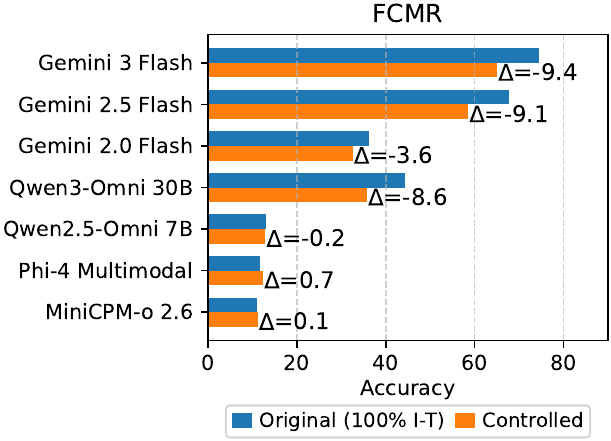}
  \caption{
Performance comparison on the \textit{original} FCMR datasets vs. their \textit{controlled} variants.
The random baseline is 12.5\%, and due to the difficulty of the dataset, most open-source models perform close to this baseline under both settings.
  }
\label{fig:CMR_IT_TI_FCMR}
\end{figure}

\section{Details of \benchmark\ Construction}
\label{sec:appendix_dataset_construction_detail}
\begin{figure*}[t]
\centering
\includegraphics[width=1.0\textwidth, keepaspectratio]{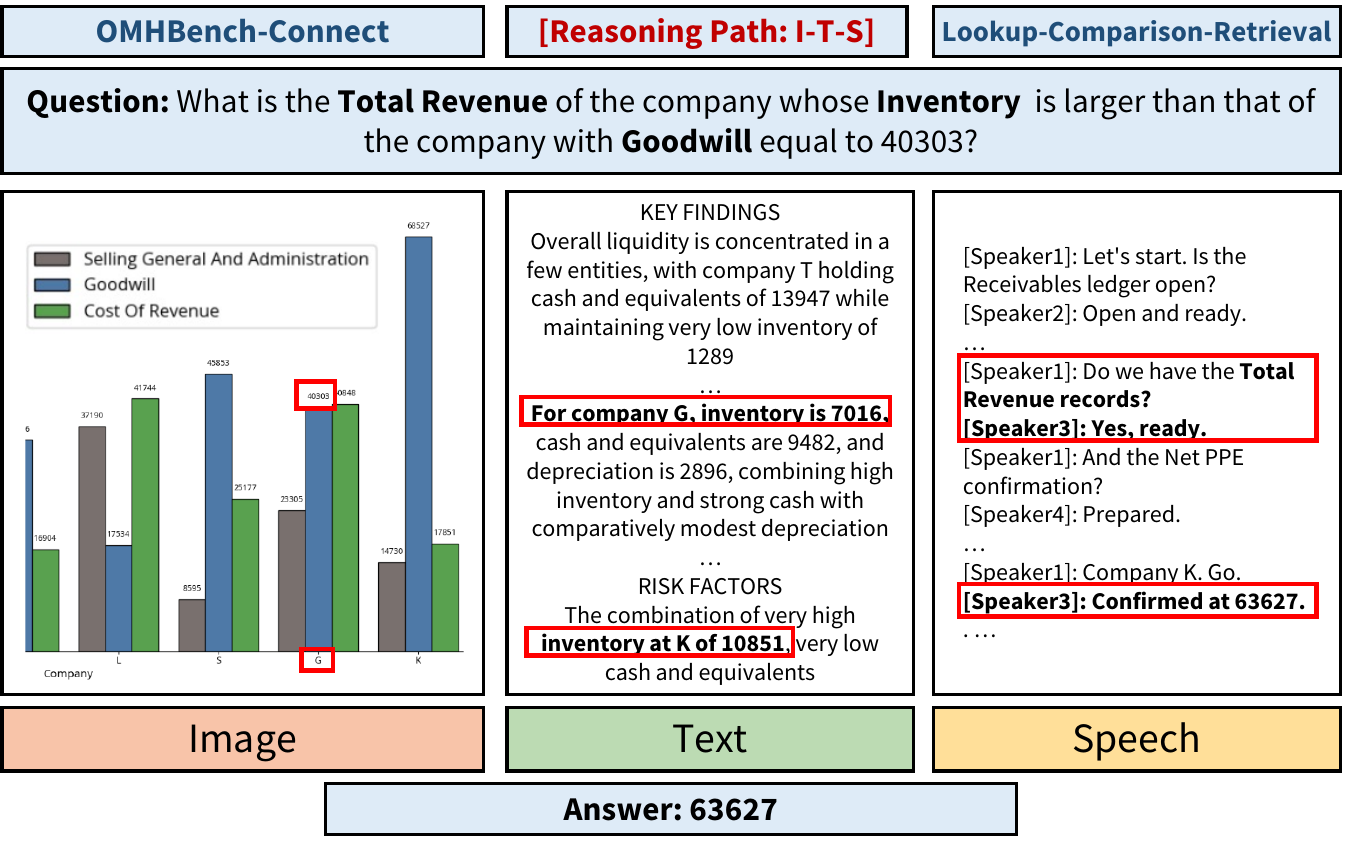}
  \caption{
Example of \benchmark-Connect I-T-S Instance.
}
\label{fig:appendix_example_its}
\end{figure*}

\begin{figure*}[t]
\centering
\includegraphics[width=1.0\textwidth, keepaspectratio]{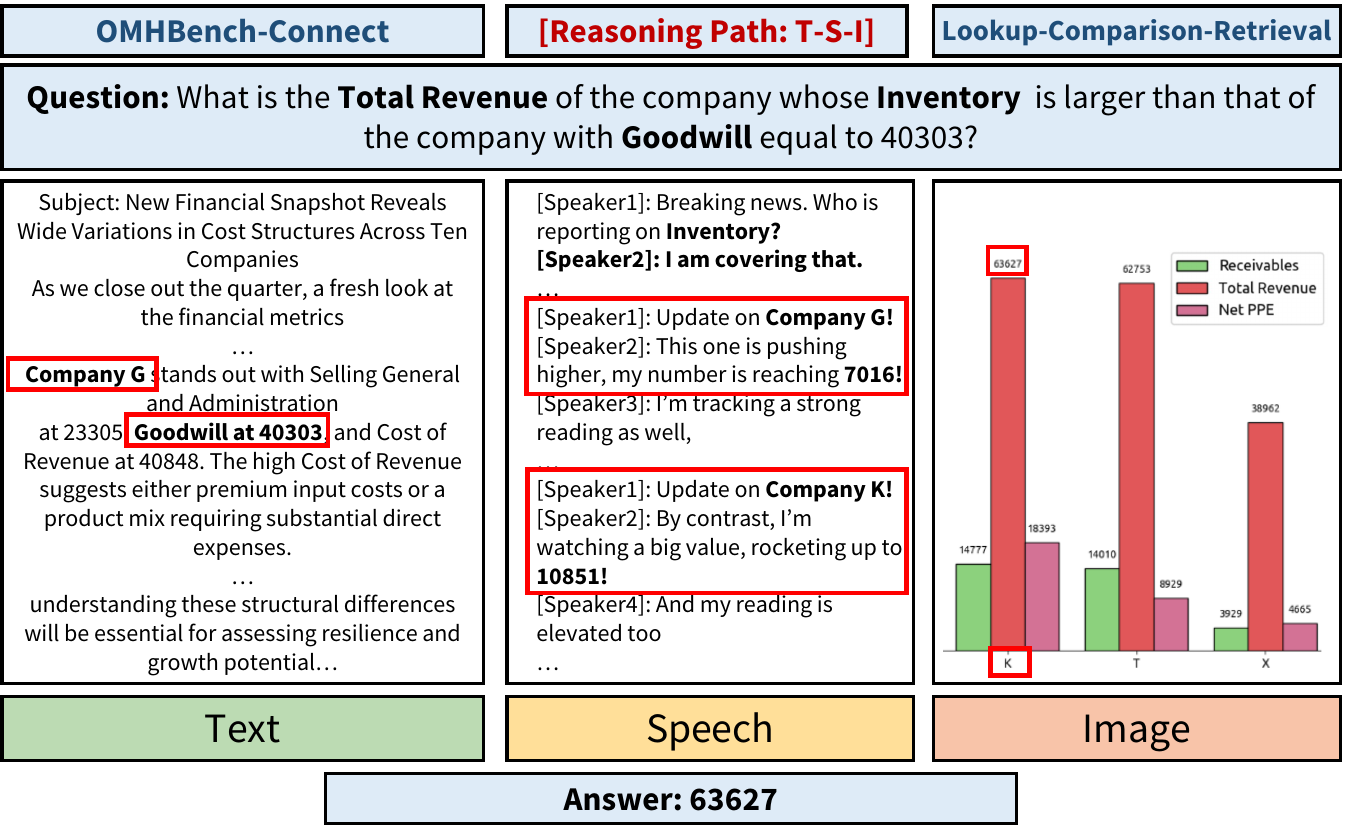}
  \caption{
Example of \benchmark-Connect T-S-I Instance.
}
\label{fig:appendix_example_tsi}
\end{figure*}

\begin{figure*}[t]
\centering
\includegraphics[width=1.0\textwidth, keepaspectratio]{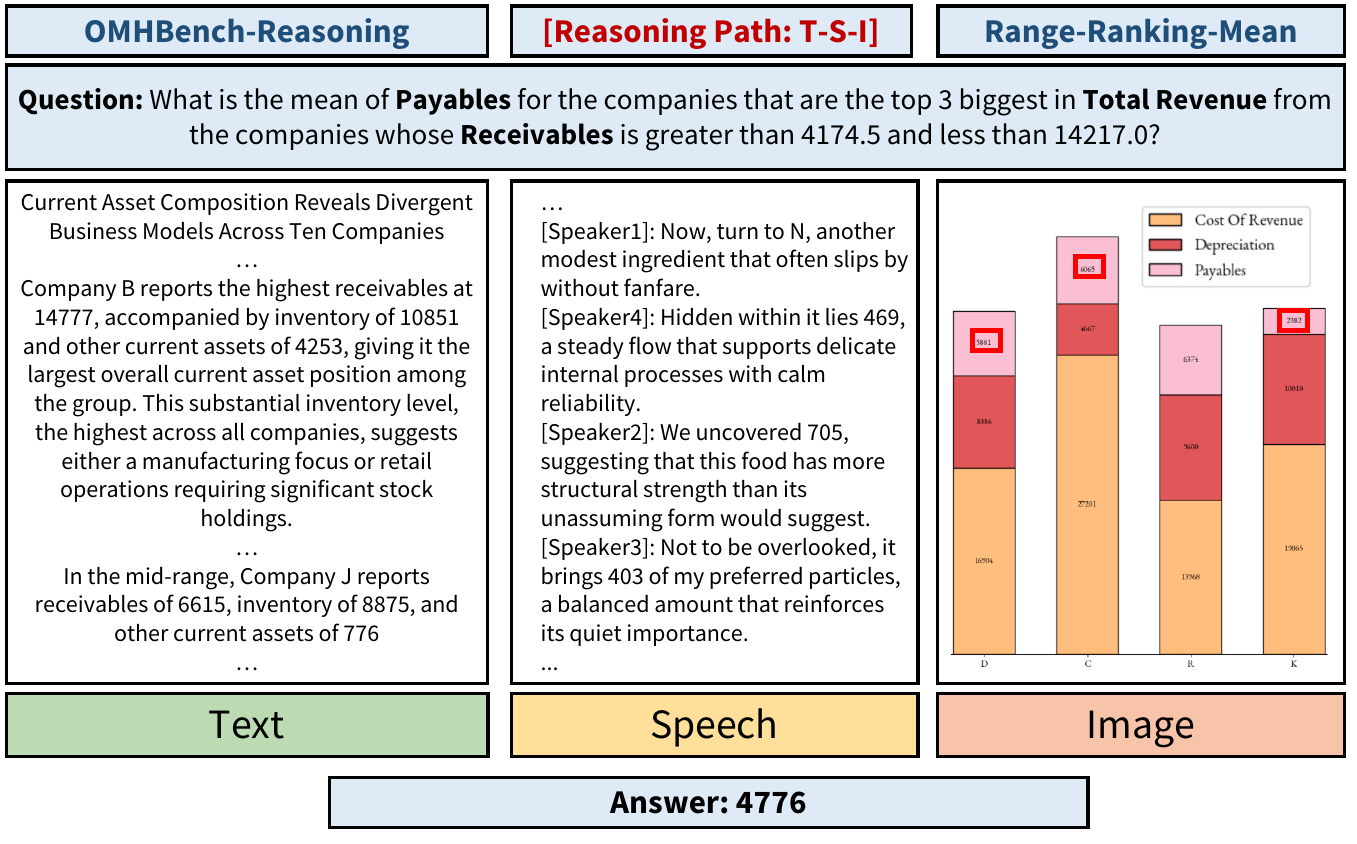}
  \caption{
Example of \benchmark-Reasoning T-S-I Instance.
}
\label{fig:appendix_example_tsi_reasoning}
\end{figure*}

The following describes the detailed design choices involved in constructing the benchmark.
Example samples from the dataset are shown in Figure \ref{fig:appendix_example_its}, Figure \ref{fig:appendix_example_tsi} and Figure \ref{fig:appendix_example_tsi_reasoning}.

\subsection{Data Source}
\label{sec:appendix_data_source}
\benchmark\ comprises four domains: finance, economics, climate, and nutrition. 
The data for each domain were obtained from Yahoo Finance\footnote{\url{https://finance.yahoo.com/}}, World Bank\footnote{\url{https://data.worldbank.org/}}, Open-Meteo\footnote{\url{https://open-meteo.com/}}, and U.S. Department of Agriculture (USDA)\footnote{\url{https://fdc.nal.usda.gov/}}, respectively.
All values are standardized to use consistent units across entities to ensure comparability.
To avoid ambiguous answers, attributes with duplicate values across samples are excluded from the dataset.

\paragraph{Finance Domain}
The finance domain is constructed using annual financial statements from 23 publicly listed companies.
The selected companies and their ticker symbols are:
MSFT, NVDA, AVGO, QCOM, TXN, IBM, ADI, MU, KLAC, LLY, MRK, ABBV, TMO, PFE, GILD, BMY, ZTS, PG, KO, PEP, DIS, MDLZ, and HON.
For each company, we extract 15 standardized financial attributes from their 2024 annual reports obtained from Yahoo Finance.
These indicators span various aspects of corporate performance, including Total Revenue, Cost Of Revenue, Selling General And Administration, Cash And Equivalents, Receivables, Inventory, Other Current Assets, Other Non Current Assets, Net PPE, Goodwill, Payables, Long Term Debt, Other Non Current Liabilities, Depreciation, and Stock Based Compensation.

\begin{table}[t]
  \centering
  \small
  \setlength{\tabcolsep}{8pt}

  \begin{tabular}{@{}l c@{}}
    \toprule
    \textbf{Statistics} & \textbf{Number} \\
    \midrule
    Dataset &  \\
    \hspace{1em}\benchmark-Connect & 3,072 \\
    \hspace{1em}\benchmark-Reasoning & 3,072 \\
    Total samples & 6,144 \\
    \midrule
    Reasoning path & 6 \\
    \hspace{1em}I-S-T & 1,024 \\
    \hspace{1em}I-T-S & 1,024 \\
    \hspace{1em}S-I-T & 1,024 \\
    \hspace{1em}S-T-I & 1,024 \\
    \hspace{1em}T-I-S & 1,024 \\
    \hspace{1em}T-S-I & 1,024 \\
    \midrule
    Domain & 4 \\
    \hspace{1em}Finance & 1,536 \\
    \hspace{1em}Economics & 1,536 \\
    \hspace{1em}Climate & 1,536 \\
    \hspace{1em}Nutrition & 1,536 \\
    \midrule
    Operation types & 8 \\
    Operation combinations & 33 \\
    \midrule
    Image type & 10 \\
    Image color & 20 \\
    Image font & 20 \\
    Plot library & 2 \\
    \midrule
    Speech type & 22 \\
    Speech voice & 27 \\
    \midrule
    Text type & 24 \\
    \midrule
    Generation LLM variants & 3 \\
    \bottomrule
  \end{tabular}
  \caption{Key statistics of \benchmark.}
  \normalsize
  \label{tab:dataset_statistics_table}
\end{table}

\paragraph{Economics Domain}
The economics domain is constructed using country-level economic attributes obtained from the World Bank World.
All values correspond to annual data for the year 2024.
The dataset includes 18 countries: ARG, AUS, BRA, CHE, DEU, EGY, ESP, FRA, GBR, IDN, IND, ITA, MEX, NLD, NOR, SAU, SWE, and ZAF
For each country, we collect 18 standardized economic indicators representing major components of national economic activity.
The selected indicators include Personal remittances, paid, Personal remittances, received, Total reserves, excluding gold, Final consumption expenditure, Gross fixed capital formation, Gross capital formation, Agriculture, forestry, and fishing, value added, Manufacturing, value added, Industry, including construction, value added, Services, value added, Gross Value Added (GVA) at basic prices, GNI, Gross savings, Taxes less subsidies on products, Merchandise imports, Commercial service imports, Merchandise exports, and Commercial service exports.

\paragraph{Climate Domain}
The climate domain is constructed using monthly meteorological data collected from major cities around the world.
All climate data correspond to the year 2024 and are obtained from Open-Meteo.
The dataset includes 20 representative cities: Karachi, Addis Ababa, Cairo, Nairobi, Los Angeles, Tokyo, Ho Chi Minh City, Ulaanbaatar, Chicago, Singapore, Toronto, Shanghai, Manila, London, Lagos, Chengdu, Beijing, Dubai, Rome, and Mumbai.
These cities were selected to cover diverse geographic regions and climate conditions.
For each city, we collect 12 climate attributes corresponding to the maximum wind speed for each month from January to December.

\paragraph{Nutrition Domain}
The nutrition domain is constructed using food composition data obtained from the U.S. Department of Agriculture (USDA).
The dataset includes 24 food items: Potatoes, mashed, dehydrated, granules without milk, dry form; Sorghum flour, whole-grain; Wheat, KAMUT khorasan, uncooked; PAPA JOHN'S 14" The Works Pizza, Original Crust; Lasagna with meat sauce, frozen, prepared; Potatoes, mashed, home-prepared, whole milk added; Frankfurter, turkey; Seeds, sesame seed kernels, dried (decorticated); Pork, cured, ham -- water added, slice, bone-in, separable lean and fat, unheated; Nuts, cashew nuts, raw; T.G.I. FRIDAY'S, chicken fingers, from kids' menu; Pork, cured, ham with natural juices, shank, bone-in, separable lean only, unheated; Broccoli, cooked, boiled, drained, with salt; Pork, cured, ham and water product, shank, bone-in, unheated, separable lean only; Bologna, beef; Teff, uncooked; Nuts, pecans; HOT POCKETS Ham 'N Cheese Stuffed Sandwich, frozen; Pork, cured, ham -- water added, slice, boneless, separable lean only, heated, pan-broil; Pork sausage, link/patty, fully cooked, microwaved; DENNY'S, chicken strips; Pork, cured, ham and water product, rump, bone-in, separable lean only, heated, roasted; Pork, cured, ham with natural juices, spiral slice, boneless, separable lean only, unheated; Kielbasa, fully cooked, unheated;
For each food item, 19 nutritional attributes are collected.
These include Ash, Protein, Lysine, Methionine, Isoleucine, Leucine, Valine, Phenylalanine, Threonine, Histidine, Arginine, Tyrosine, Alanine, Glycine, Serine, Proline, Tryptophan, Cystine, and Glucose.



\subsection{Table Triplet Formation}
In the first stage of the dataset generation framework, namely the Table Triplet Formation stage, we construct table triplets consisting of three tables that share the same set of entities but contain different attributes.
Each table contains 10 entities and 3 attributes, resulting in a table size of $10 \times 3$.

When forming each table triplet, we ensure that the selected attributes are distinct across the three tables while referring to the same set of entities.
To minimize information redundancy, we further curate the attributes so that they are mutually independent and not logically or algebraically derived from one another.
In particular, we exclude derived variables such as accounting identities, ratios, and other linearly dependent quantities (e.g., relationships of the form $a_1 + a_2 = a_3$), and retain only primitive, non-derived attributes.
In addition, to facilitate subsequent conversion into chart-based representations, we constrain the value ranges of the attributes such that the ratio between the maximum and minimum values does not exceed 30.
This restriction prevents extreme scale differences across attributes and ensures stable visualization and comparison across tables.

\begin{table}[t]
\centering
\small
\setlength{\tabcolsep}{6pt}
\begin{tabular}{lc}
\hline
\textbf{Operation sequence} & \textbf{\# Instances} \\
\hline
\multicolumn{2}{l}{\textit{Connect}} \\
Lookup--Comparison--Retrieval & 3,072 \\
\hline
\multicolumn{2}{l}{\textit{Reasoning}} \\
Ranking--Ranking--Mean & 96 \\
Ranking--Ranking--Summation & 96 \\
Ranking--Range--Mean & 96 \\
Ranking--Range--Summation & 96 \\
Ranking--Comparison--Mean & 96 \\
Ranking--Comparison--Summation & 96 \\
Ranking--Proximity--Mean & 96 \\
Ranking--Proximity--Summation & 96 \\
Range--Ranking--Mean & 96 \\
Range--Ranking--Summation & 96 \\
Range--Range--Mean & 96 \\
Range--Range--Summation & 96 \\
Range--Comparison--Mean & 96 \\
Range--Comparison--Summation & 96 \\
Range--Proximity--Mean & 96 \\
Range--Proximity--Summation & 96 \\
Comparison--Ranking--Mean & 96 \\
Comparison--Ranking--Summation & 96 \\
Comparison--Range--Mean & 96 \\
Comparison--Range--Summation & 96 \\
Comparison--Comparison--Mean & 96 \\
Comparison--Comparison--Summation & 96 \\
Comparison--Proximity--Mean & 96 \\
Comparison--Proximity--Summation & 96 \\
Proximity--Ranking--Mean & 96 \\
Proximity--Ranking--Summation & 96 \\
Proximity--Range--Mean & 96 \\
Proximity--Range--Summation & 96 \\
Proximity--Comparison--Mean & 96 \\
Proximity--Comparison--Summation & 96 \\
Proximity--Proximity--Mean & 96 \\
Proximity--Proximity--Summation & 96 \\
\hline
\end{tabular}
\caption{Operation sequences used to construct \benchmark-Connect and \benchmark-Reasoning.}
\label{tab:appendix_operation_combination}
\end{table}

\subsection{Multi-Hop QA Construction}
\label{sec:appendix_dataset_construction_mhqa_const}
\benchmark\ constructs each question as a three-hop reasoning chain over entities and their attributes. 
The first two hops select or filter sets of entities, while the final hop produces a scalar numerical answer.
Each hop applies a specific operation to a designated attribute, and all operations are deterministic and rule-based, ensuring full control over the reasoning path.
Below, we describe the concrete mechanics of each operation in detail.
The valid sequences of operations used to construct multi-hop questions are summarized in Table \ref{tab:appendix_operation_combination}.

\paragraph{\textbf{Lookup}}
Lookup anchors the reasoning chain to a single entity.
From a base table containing ten entities, one entity is randomly selected, and the value of a specified attribute is retrieved.
A strict uniqueness constraint is enforced: the selected attribute value must occur exactly once in the table.
If the same value appears for multiple entities, the instance is discarded.
The output of this operation is a uniquely identified entity and its corresponding attribute value.

\paragraph{\textbf{Comparison}}
Comparison filters an entity set using a strict inequality condition on a specified attribute.
Entities are sorted by the attribute, and a threshold value is constructed such that exactly a predefined number of entities satisfy either a ``larger than'' or ``smaller than'' condition.
Instances in which ties occur at the decision boundary are discarded.
This operation outputs a reduced entity set of fixed size.

\paragraph{\textbf{Ranking}}
Ranking selects entities based on their ordinal position under a given attribute.
Entities are sorted in ascending or descending order, and a fixed number of top or bottom entities are selected.
No explicit threshold values are involved.
The output is a subset of entities.

\paragraph{\textbf{Range}}
Range selects entities whose attribute values fall within a contiguous interval.
Entities are first sorted by the target attribute, and a consecutive segment is selected.
The interval boundaries are defined to ensure that the selected entities are uniquely determined.
The output is a subset of entities.

\paragraph{\textbf{Proximity}}
Proximity selects entities whose attribute values are closest to a reference value.
Entities are ranked by their absolute distance to the reference, and the closest entities are selected.
The output is a subset of entities.

\paragraph{\textbf{Retrieval}}
Retrieval is applied to a uniquely determined entity.
Given a target attribute, the corresponding attribute value is retrieved from the table and returned as the final answer.

\paragraph{\textbf{Summation}}
Summation aggregates a numerical attribute over a filtered entity set by summing all corresponding values.
The output is a scalar numerical value.

\paragraph{\textbf{Mean}}
Mean computes the arithmetic average of a numerical attribute over a filtered entity set.
Instances in which the resulting mean is non-integer are discarded during dataset construction.
The output is a scalar numerical value.

\subsection{Omni-Modal Context Generation}

\paragraph{Image Modality}
We convert tabular data into image modality using widely adopted visualization libraries, Matplotlib and Seaborn. 
A total of ten chart types are generated: vertical bar, horizontal bar, vertical stacked bar, horizontal stacked bar, lollipop, line, scatter, heatmap, bubble, and tile.

To enhance visual diversity, we randomly select one of 20 fonts for each image, uniformly sampled across all images.
The fonts used are: Arimo[wght], FiraSansCondensed-Regular, OpenSans-Regular, RobotoSlab[wght], WorkSans-VariableFont[wght], CALIBRI, Kosugi-Regular, OpenSansHebrew-Regular, SourceSansPro-Regular, arial, EBGaramond-VariableFont[wght], Lato-Regular, OpenSansHebrewCondensed-Regular, Tinos-Regular, tahoma, FiraSans-Regular, NotoSans-Regular, Roboto-Regular, Ubuntu-Regular, and times.

In addition, chart elements are colored using a fixed palette of 20 distinct colors: \#4E79A7, \#A0CBE8, \#F28E2B, \#FFBE7D, \#59A14F, \#8CD17D, \#B6992D, \#F1CE63, \#499894, \#86BCB6, \#E15759, \#FF9D9A, \#79706E, \#BAB0AC, \#D37295, \#FABFD2, \#B07AA1, \#D4A6C8, \#9D7660, and \#D7B5A6.
These design choices allow us to construct a rich and diverse set of images.

\paragraph{Text Modality}
We define a total of 24 representative text scenarios across four domains.
For the finance domain, the scenarios include Analyst Report, News Article, Blog Post, Email Newsletter, Executive Summary, and Meeting Minutes.
For the economics domain, the scenarios include Analyst Report, News Article, Blog Post, Email Newsletter, Executive Summary, and Meeting Minutes.
For the climate domain, the scenarios include Research Report, Business Report, City Marketing, News Article, Blog Post, and Magazine.
For the nutrition domain, the scenarios include Research Report, Quality Assurance Log, Dietary Guidelines, Ingredient Encyclopedia, Blog Post, and Magazine.

\begin{figure*}[t]
\centering
\includegraphics[width=1.0\textwidth, keepaspectratio]{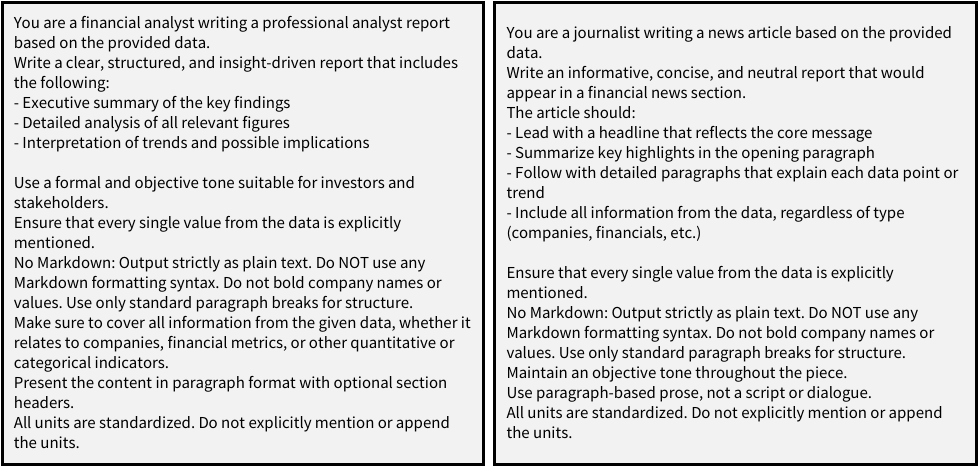}
  \caption{
Example prompts used to generate the text.
}
\label{fig:appendix_text_modality_prompt_finance}
\end{figure*}

For each scenario, we design a tailored situational prompt and use large language models to convert structured tabular data into natural language text.
Figure \ref{fig:appendix_text_modality_prompt_finance} illustrates an example of the prompt used for text scenario generation.

\begin{figure*}[t]
\centering
\includegraphics[width=1.0\textwidth, keepaspectratio]{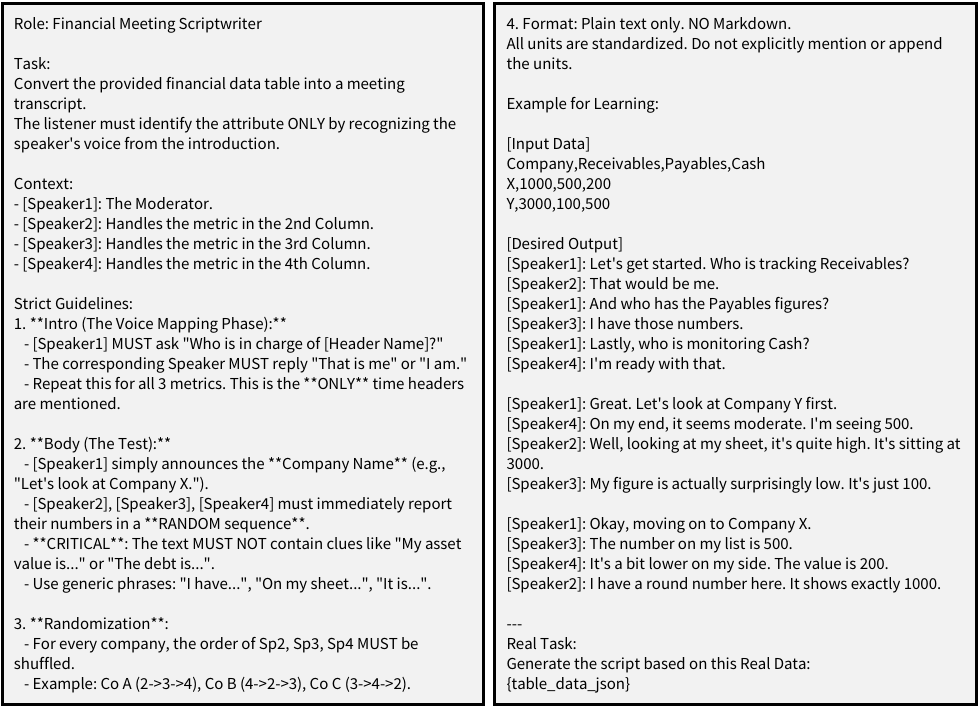}
  \caption{
Example prompt used to generate the speech script.
}
\label{fig:appendix_speech_modality_prompt_finance}
\end{figure*}

\paragraph{Speech Modality}
We define 22 representative speech scenarios across four domains.
For the finance domain, the scenarios include Meeting, Podcast, Seminar, Audit, and News Debate.
For the economics domain, the scenarios include Meeting, Podcast, Seminar, News Debate, and Global Summit.
For the climate domain, the scenarios include Weather Forecast, Meeting, Airport Control Tower, Sports Event Briefing, Business Risk Briefing, and Green Energy Assessment.
For the nutrition domain, the scenarios include Meeting, Lab Briefing, Documentary, Ingredient Safety Audit, Conference, and Podcast.
Each scenario consists of a four-speaker dialogue.
One speaker serves as the moderator, while the remaining three speakers are each responsible for different attributes.
Similar to the text modality, we first use Large Language Models to convert structured tabular data into textual scripts tailored to each scenario.
These scripts are then converted into speech using the Kokoro-82M TTS model \cite{hexgrad_2025}.
To enhance acoustic diversity, we utilize 27 distinct voices provided by the Kokoro-82M model, including 14 female and 13 male voices.\footnote{The full list of available voices is provided at \url{https://huggingface.co/hexgrad/Kokoro-82M/blob/main/VOICES.md}}
The female voices include:
af\_heart, af\_alloy, af\_aoede, af\_bella, af\_jessica, af\_kore, af\_nova, af\_river, af\_sarah, af\_sky, bf\_alice, bf\_emma, bf\_isabella, bf\_lily.
The male voices include:
am\_adam, am\_echo, am\_eric, am\_fenrir, am\_liam, am\_michael, am\_onyx, am\_puck, am\_santa, bm\_daniel, bm\_fable, bm\_george, bm\_lewis.
This voice configuration ensures a high degree of variation in the generated speech data.
Figure \ref{fig:appendix_speech_modality_prompt_finance} illustrates an example of the prompt used for speech scenario generation.

\section{Comparison of Evaluation Methods}
\label{app:eval-method-comparison}

We further validate Exact Match in \benchmark\ by comparing it with human judgment and LLM-as-a-Judge evaluation. Specifically, we randomly sampled 300 instances from \benchmark-Connect and 300 instances from \benchmark-Reasoning, and evaluated Gemini 3 Flash outputs under all three protocols. All methods yielded identical results: 72.7\% on \benchmark-Connect and 45.7\% on \benchmark-Reasoning. This agreement was consistent across multiple judge models, including Gemini 3 Flash \cite{gemini3Flash_modelcard}, GPT 5.2 \cite{openai_gpt5p2_systemcard}, Claude Haiku 4.5 \cite{anthropic_claude_haiku45_systemcard}, and Qwen3-235B-A22B \cite{yang2025qwen3}.

This result is expected given the design of \benchmark. Since all ground-truth answers are positive integers, evaluation is not affected by paraphrasing or alternative surface forms. In practice, correctness reduces to checking whether the predicted final answer matches the target numeric value, making Exact Match fully consistent with both human and LLM-based evaluation in this benchmark.

\section{Path Balance Score at Different Thresholds}
\label{app:pbs_at_k}

To provide a more fine-grained view of robustness to reasoning-path variation, we additionally report \textbf{PBS@k}. While PBS measures whether a model answers all reasoning-path variants of the same question correctly, PBS@k measures the proportion of question groups answered correctly on at least $k$ out of the six paths.

Formally, let the dataset contain $|D|$ instances, forming $|G| = |D|/N!$ groups. For the $i$-th group, let $a_{i,j} \in \{0,1\}$ denote whether the model correctly answers the $j$-th path. PBS@k is defined as:
\[
\scalebox{0.95}{$
\text{PBS@}k = \frac{1}{|G|}
\sum_{i=1}^{|G|}
\mathbb{I}\left(\sum_{j=1}^{N!} a_{i,j} \ge k\right),
$}
\]
where $\mathbb{I}(\cdot)$ is the indicator function and $k \in \{1,\dots,N!\}$. In \benchmark, $N=3$, so each question has six reasoning-path variants. Thus, PBS@1 indicates whether a model succeeds on at least one path variant, while PBS@6 coincides with PBS. Tables~\ref{tab:pbs_at_k_connect} and~\ref{tab:pbs_at_k_reasoning} report PBS@k results on \benchmark-Connect and \benchmark-Reasoning, respectively.

Across models, PBS@k decreases monotonically as $k$ increases, showing that consistency across reasoning paths varies substantially by model and benchmark split. This trend is especially pronounced in \benchmark-Reasoning, where even the strongest models exhibit a large gap between PBS@1 and PBS@6.

\begin{table}[t]
  \centering
  \scriptsize
  \setlength{\tabcolsep}{2.0pt}
  \renewcommand{\arraystretch}{1.1}
  \begin{tabularx}{\columnwidth}{
    @{}
    >{\raggedright\arraybackslash}X
    @{}
    *{6}{@{\hspace{1.2pt}}r@{\hspace{1.2pt}}}
    @{}
  }
    \toprule
    \multirow{2}{*}[-0.9ex]{\textbf{Model}} &
    \multicolumn{6}{c}{\textbf{PBS@k (\%)}} \\
    \cmidrule(lr){2-7}
    & \textbf{$k=1$} & \textbf{$k=2$} & \textbf{$k=3$} &
      \textbf{$k=4$} & \textbf{$k=5$} & \textbf{$k=6$} \\
    \midrule
    \textbf{\textit{Proprietary Models}} \\
    Gemini 3 Flash        & 100.0 & 99.6 & 95.5 & 83.8 & 58.8 & 32.2 \\
    Gemini 2.5 Pro        & 100.0 & 99.2 & 94.5 & 72.7 & 43.8 & 25.0 \\
    Gemini 2.5 Flash      & 96.9  & 91.4 & 70.3 & 43.0 & 15.6 & 4.7 \\
    Gemini 2.5 Flash-lite & 87.5  & 61.7 & 30.5 & 13.3 & 0.8  & 0.0 \\
    Gemini 2.0 Flash      & 73.4  & 39.8 & 14.1 & 1.6  & 0.8  & 0.0 \\
    Gemini 2.0 Flash-lite & 72.2  & 28.6 & 6.3  & 0.0  & 0.0  & 0.0 \\
    \midrule
    \textbf{\textit{Open-Source Models}} \\
    Qwen3-Omni 30B        & 97.9  & 85.9 & 55.7 & 31.2 & 8.0  & 2.3 \\
    Phi-4 Multimodal      & 58.8  & 24.4 & 6.2  & 1.2  & 0.0  & 0.0 \\
    Qwen2.5-Omni 7B       & 57.8  & 22.5 & 5.7  & 1.2  & 0.0  & 0.0 \\
    Qwen2.5-Omni 3B       & 46.1  & 14.6 & 2.5  & 0.4  & 0.0  & 0.0 \\
    OmniVinci             & 35.0  & 9.4  & 2.1  & 0.2  & 0.0  & 0.0 \\
    MiniCPM-o 2.6         & 30.3  & 5.1  & 0.8  & 0.0  & 0.0  & 0.0 \\
    Omni-AutoThink        & 24.6  & 4.1  & 0.2  & 0.0  & 0.0  & 0.0 \\
    \bottomrule
  \end{tabularx}
  \caption{
PBS@k results on \textbf{\benchmark-Connect}. PBS@k measures the proportion of question groups answered correctly on at least $k$ of the six reasoning paths. PBS@6 coincides with PBS.
}
  \label{tab:pbs_at_k_connect}
\end{table}

\begin{table}[t]
  \centering
  \scriptsize
  \setlength{\tabcolsep}{2.0pt}
  \renewcommand{\arraystretch}{1.1}
  \begin{tabularx}{\columnwidth}{
    @{}
    >{\raggedright\arraybackslash}X
    @{}
    *{6}{@{\hspace{1.2pt}}r@{\hspace{1.2pt}}}
    @{}
  }
    \toprule
    \multirow{2}{*}[-0.9ex]{\textbf{Model}} &
    \multicolumn{6}{c}{\textbf{PBS@k (\%)}} \\
    \cmidrule(lr){2-7}
    & \textbf{$k=1$} & \textbf{$k=2$} & \textbf{$k=3$} &
      \textbf{$k=4$} & \textbf{$k=5$} & \textbf{$k=6$} \\
    \midrule
    \textbf{\textit{Proprietary Models}} \\
    Gemini 3 Flash        & 90.0 & 77.5 & 57.2 & 41.2 & 22.1 & 8.6 \\
    Gemini 2.5 Pro        & 91.4 & 76.6 & 56.2 & 36.7 & 21.1 & 10.9 \\
    Gemini 2.5 Flash      & 67.2 & 36.7 & 14.1 & 6.2  & 1.6  & 0.0 \\
    Gemini 2.5 Flash-lite & 44.5 & 14.8 & 4.7  & 0.0  & 0.0  & 0.0 \\
    Gemini 2.0 Flash      & 24.2 & 3.1  & 0.8  & 0.0  & 0.0  & 0.0 \\
    Gemini 2.0 Flash-lite & 14.3 & 2.4  & 0.0  & 0.0  & 0.0  & 0.0 \\
    \midrule
    \textbf{\textit{Open-Source Models}} \\
    Qwen3-Omni 30B        & 54.5 & 26.4 & 7.2  & 1.8  & 0.2  & 0.0 \\
    Phi-4 Multimodal      & 1.4  & 0.2  & 0.0  & 0.0  & 0.0  & 0.0 \\
    Qwen2.5-Omni 7B       & 3.1  & 1.0  & 0.2  & 0.0  & 0.0  & 0.0 \\
    Qwen2.5-Omni 3B       & 2.1  & 0.0  & 0.0  & 0.0  & 0.0  & 0.0 \\
    OmniVinci             & 1.4  & 0.0  & 0.0  & 0.0  & 0.0  & 0.0 \\
    MiniCPM-o 2.6         & 0.4  & 0.0  & 0.0  & 0.0  & 0.0  & 0.0 \\
    Omni-AutoThink        & 1.2  & 0.0  & 0.0  & 0.0  & 0.0  & 0.0 \\
    \bottomrule
  \end{tabularx}
  \caption{
PBS@k results on \textbf{\benchmark-Reasoning}. PBS@k measures the proportion of question groups answered correctly on at least $k$ of the six reasoning paths. PBS@6 coincides with PBS.
}
  \label{tab:pbs_at_k_reasoning}
\end{table}

\section{Experimental Environment}
All experiments were conducted on a machine equipped with Intel Xeon Gold 6338 CPU (2.00 GHz), and an NVIDIA A100-SXM4 GPU with 80 GB of memory.
The system ran Ubuntu 22.04.4 LTS with CUDA compilation tools release 12.4.
We used Python 3.10.18 and PyTorch 2.6.0+cu124 as the core software environment.
During both dataset generation and evaluation, the random seed was fixed to 42 to ensure reproducibility.

\begin{table*}[t]
  \centering
  \normalsize
  \setlength{\tabcolsep}{6.0pt} 
  \renewcommand{\arraystretch}{1.2}

  \begin{tabularx}{\textwidth}{
    >{\raggedright\arraybackslash}X
    *{6}{r}
    r
    r
  }
    \toprule
    \multirow{2}{*}[-0.9ex]{\textbf{Model}} &
    \multicolumn{6}{c}{\textbf{Accuracy by Reasoning Path (\%)}} &
    \multirow{2}{*}[-0.9ex]{\makecell{\textbf{Avg.}\\\textbf{(Acc.)}}} &
    \multirow{2}{*}[-0.9ex]{\textbf{PBS}} \\
    \cmidrule(lr){2-7}
    & \textbf{S-I-T} & \textbf{S-T-I} & \textbf{I-S-T} &
      \textbf{T-S-I} & \textbf{I-T-S} & \textbf{T-I-S} & & \\
    \midrule

    \textbf{\textit{Proprietary Models}} \\ 
    Gemini 3 Flash (Think)        & 97.5 & 98.4 & 75.4 & 75.0 & 60.2 & 63.5 & 78.3 & 32.2 \\
    Gemini 3 Flash (Non-Think)    & 93.0 & 94.5 & 60.9 & 57.8 & 43.0 & 39.1 & 64.7 & 8.6 \\
    Gemini 2.5 Pro (Think)        & 94.5 & 96.9 & 66.4 & 71.1 & 55.5 & 50.8 & 72.5 & 25.0 \\
    Gemini 2.5 Pro (Non-Think)    & 65.6 & 68.8 & 50.0 & 58.6 & 41.4 & 38.3 & 53.8 & 6.2 \\
    Gemini 2.5 Flash (Think)      & 82.0 & 85.9 & 50.8 & 54.7 & 26.6 & 21.9 & 53.6 & 4.7 \\
    Gemini 2.5 Flash (Non-Think)  & 65.6 & 69.5 & 37.5 & 39.1 & 24.2 & 15.6 & 41.9 & 2.3 \\
    Gemini 2.5 Flash-lite (Think) & 49.2 & 60.9 & 38.3 & 35.2 & 5.5  & 4.7  & 32.3 & 0.0 \\
    Gemini 2.5 Flash-lite (Non-Think)
                                  & 32.8 & 37.5 & 28.9 & 26.6 & 2.3  & 0.8  & 21.5 & 0.0 \\
    Gemini 2.0 Flash              & 28.9 & 33.6 & 26.6 & 29.7 & 4.7  & 6.2  & 21.6 & 0.0 \\
    Gemini 2.0 Flash-lite         & 35.9 & 32.8 & 21.1 & 11.7 & 2.3  & 2.3  & 17.7 & 0.0 \\

    \midrule
    \textbf{\textit{Open-Source Models}} \\ 
    Qwen3-Omni 30B (Think)        & 75.8 & 77.0 & 46.7 & 49.6 & 16.0 & 16.0 & 46.8 & 2.3 \\
    Qwen3-Omni 30B (Non-Think)    & 35.0 & 44.7 & 17.8 & 33.8 & 6.2  & 7.2  & 24.1 & 0.0 \\
    Phi-4 Multimodal              & 26.6 & 23.6 & 21.5 & 18.4 & 0.6  & 0.0  & 15.1 & 0.0 \\
    Qwen2.5-Omni 7B               & 22.7 & 20.9 & 19.3 & 20.5 & 2.0  & 1.8  & 14.5 & 0.0 \\
    Qwen2.5-Omni 3B               & 12.7 & 17.6 & 15.6 & 14.6 & 1.2  & 2.0  & 10.6 & 0.0 \\
    OmniVinci                     & 14.8 & 8.6  & 14.8 & 7.0  & 0.8  & 0.6  & 7.8  & 0.0 \\
    MiniCPM-o 2.6                 & 8.0  & 10.9 & 7.4  & 8.4  & 1.2  & 0.2  & 6.0  & 0.0 \\
    Omni-AutoThink                & 7.6  & 6.6  & 8.0  & 6.1  & 0.6  & 0.0  & 4.8  & 0.0 \\

    \bottomrule
  \end{tabularx}

  \caption{
  Accuracies and Path Balance Scores (PBSs) across six reasoning paths in \textbf{\benchmark-Connect}, including both thinking and non-thinking variants.
  Avg denotes macro-averaged accuracy.
  PBSs measure robustness to reasoning path variations.
  }
  \label{tab:accuracy_by_reasoning_path_Connect_full_model}
\end{table*}

\begin{table*}[t]
  \centering
  \normalsize
  \setlength{\tabcolsep}{6.0pt} 
  \renewcommand{\arraystretch}{1.2}

  \begin{tabularx}{\textwidth}{
    >{\raggedright\arraybackslash}X
    *{6}{r}
    r
    r
  }
    \toprule
    \multirow{2}{*}[-0.9ex]{\textbf{Model}} &
    \multicolumn{6}{c}{\textbf{Accuracy by Reasoning Path (\%)}} &
    \multirow{2}{*}[-0.9ex]{\makecell{\textbf{Avg.}\\\textbf{(Acc.)}}} &
    \multirow{2}{*}[-0.9ex]{\textbf{PBS}} \\
    \cmidrule(lr){2-7}
    & \textbf{S-I-T} & \textbf{S-T-I} & \textbf{I-S-T} &
      \textbf{T-S-I} & \textbf{I-T-S} & \textbf{T-I-S} & & \\
    \midrule

    \textbf{\textit{Proprietary Models}} \\ 
    Gemini 3 Flash (Think)        & 55.9 & 58.8 & 49.8 & 49.6 & 40.0 & 42.6 & 49.4 & 8.6 \\
    Gemini 3 Flash (Non-Think)    & 35.9 & 39.8 & 31.2 & 29.7 & 16.4 & 18.8 & 28.6 & 1.6 \\
    Gemini 2.5 Pro (Think)        & 53.9 & 51.6 & 52.3 & 47.7 & 41.4 & 46.1 & 48.8 & 10.9 \\
    Gemini 2.5 Pro (Non-Think)    & 28.9 & 32.8 & 30.5 & 28.9 & 14.8 & 21.1 & 26.2 & 0.0 \\
    Gemini 2.5 Flash (Think)      & 32.0 & 30.5 & 17.2 & 24.2 & 10.9 & 10.9 & 21.0 & 0.0 \\
    Gemini 2.5 Flash (Non-Think)  & 22.7 & 24.2 & 17.2 & 14.8 & 6.2  & 5.5  & 15.1 & 0.0 \\
    Gemini 2.5 Flash-lite (Think) & 18.8 & 21.1 & 15.6 & 8.6  & 0.0  & 0.0  & 10.7 & 0.0 \\
    Gemini 2.5 Flash-lite (Non-Think)
                                  & 7.8  & 9.4  & 3.9  & 5.5  & 0.0  & 0.8  & 4.6  & 0.0 \\
    Gemini 2.0 Flash              & 4.7  & 11.7 & 4.7  & 6.2  & 0.8  & 0.0  & 4.7  & 0.0 \\
    Gemini 2.0 Flash-lite         & 3.9  & 5.5  & 3.9  & 2.3  & 0.8  & 0.0  & 2.7  & 0.0 \\

    \midrule
    \textbf{\textit{Open-Source Models}} \\ 
    Qwen3-Omni 30B (Think)        & 27.3 & 28.5 & 14.1 & 14.6 & 2.7 & 2.7 & 15.0 & 0.0 \\
    Qwen3-Omni 30B (Non-Think)    & 9.8  & 10.5 & 4.7  & 6.4  & 0.8 & 0.6 & 5.5  & 0.0 \\
    Phi-4 Multimodal              & 0.6  & 0.4  & 0.2  & 0.0  & 0.2 & 0.2 & 0.3  & 0.0 \\
    Qwen2.5-Omni 7B               & 0.4  & 1.0  & 1.0  & 0.6  & 0.2 & 1.2 & 0.7  & 0.0 \\
    Qwen2.5-Omni 3B               & 0.8  & 0.6  & 0.2  & 0.0  & 0.4 & 0.2 & 0.4  & 0.0 \\
    OmniVinci                     & 0.6  & 0.2  & 0.2  & 0.4  & 0.0 & 0.0 & 0.2  & 0.0 \\
    MiniCPM-o 2.6                 & 0.0  & 0.0  & 0.0  & 0.0  & 0.0 & 0.0 & 0.0  & 0.0 \\
    Omni-AutoThink                & 0.4  & 0.2  & 0.4  & 0.2  & 0.0 & 0.0 & 0.2  & 0.0 \\

    \bottomrule
  \end{tabularx}

  \caption{
  Accuracies and Path Balance Scores (PBSs) across six reasoning paths in \textbf{\benchmark-Reasoning}, including both thinking and non-thinking variants.
  Avg denotes macro-averaged accuracy.
  PBSs measure robustness to reasoning path variations.
  }
  \label{tab:accuracy_by_reasoning_path_Reasoning_full_model}
\end{table*}

\begin{figure*}[t]
\centering
\includegraphics[width=0.95\textwidth, keepaspectratio]{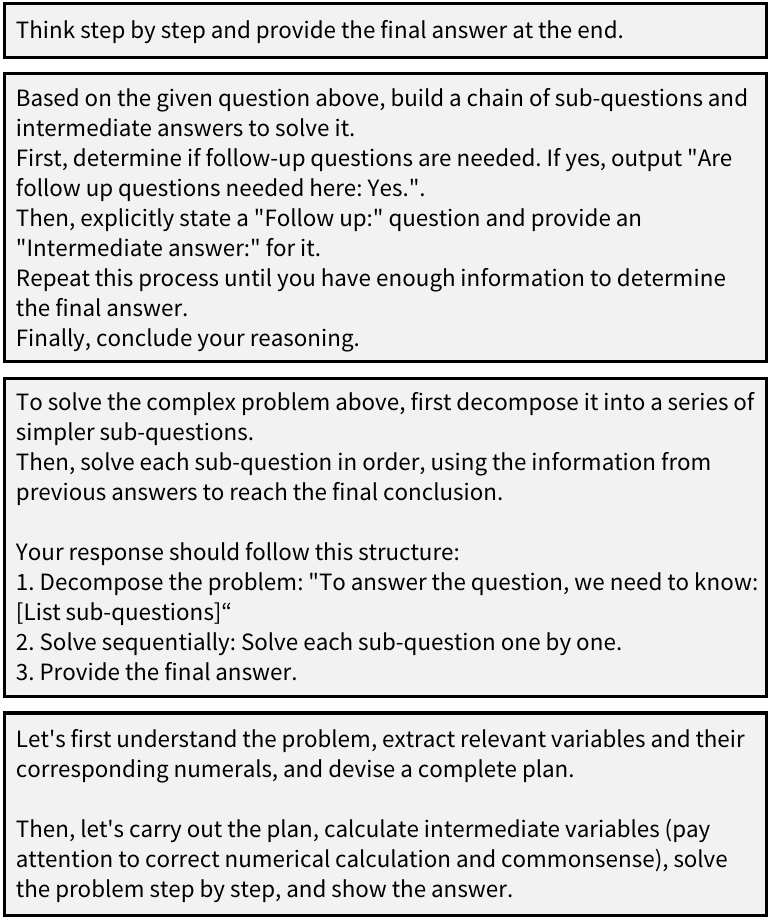}
\caption{
Prompt formats used for each prompting method: Zero-Shot CoT, Self-Ask, Least-to-Most, and Plan-and-Solve.
}
\label{fig:appendix_zero_shot_cot_prompt}
\end{figure*}

\begin{figure*}[t]
\centering
\includegraphics[width=1.0\textwidth, keepaspectratio]{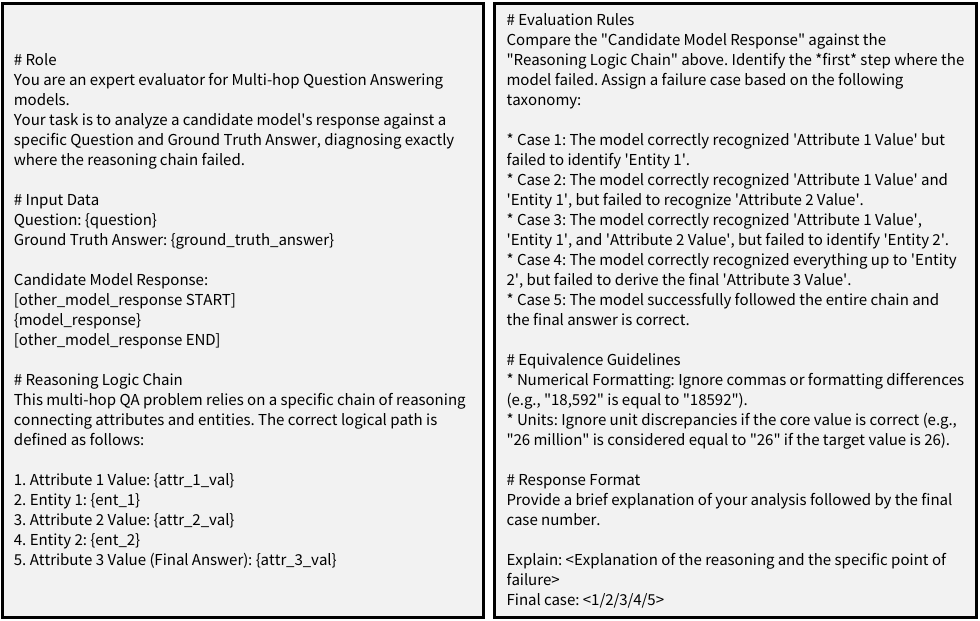}
  \caption{
Prompt used to Step-by-Step Failure Analysis.
}
\label{fig:appendix_step_by_step_analysis}
\end{figure*}

\end{document}